\documentclass[pdflatex,sn-mathphys-num,iicol]{sn-jnl}%

\usepackage{graphicx}%
\usepackage{multirow}%
\usepackage{amsmath,amssymb,amsfonts}%
\usepackage{amsthm}%
\usepackage[title]{appendix}%
\usepackage{xcolor}%
\usepackage{textcomp}%
\usepackage{manyfoot}%
\usepackage{booktabs}%
\usepackage{algorithm}%
\usepackage{algorithmicx}%
\usepackage{algpseudocode}%
\usepackage{listings}%
\usepackage{tabularx}
\usepackage{lettrine}
\usepackage{wrapfig}

\theoremstyle{thmstyleone}%

\theoremstyle{thmstyletwo}%

\theoremstyle{thmstylethree}%

\begin{document}

\title[Article Title]{DriveSplat: Unified Neural Gaussian Reconstruction for Dynamic Driving Scenes}

\author[1,2]{\fnm{Cong} \sur{Wang}}

\author[1,3]{\fnm{Ruiqi} \sur{Song}}

\author[4]{\fnm{Wei} \sur{Tian}}

\author[5]{\fnm{Chenming} \sur{Zhang}}

\author[6]{\fnm{Lingxi} \sur{Li}}

\author*[1,3]{\fnm{Long} \sur{Chen}}\email{long.chen@ia.ac.cn}

\affil[1]{\orgdiv{the State Key Laboratory of Multimodal Artificial Intelligence Systems}, \orgname{Institute of Automation, Chinese Academy of Sciences}, \orgaddress{\city{Beijing}, \country{China}}}
\affil[2]{\orgdiv{Zhongguancun Academy}, \orgaddress{\city{Beijing}, \country{China}}}
\affil[3]{\orgdiv{Waytous}, \orgaddress{\city{Beijing}, \country{China}}}
\affil[4]{\orgdiv{School of Automotive Studies}, \orgname{Tongji University}, \orgaddress{\city{Shanghai}, \country{China}}}
\affil[5]{\orgdiv{Institute of Artificial Intelligence and Robotics},  \orgname{Xi'an Jiaotong University}, \orgaddress{\city{Xi'an}, \country{China}}}
\affil[6]{\orgdiv{School of Electrical and Computer Engineering}, \orgname{Purdue University}, \orgaddress{\city{Indianapolis}, \country{USA}}}

\abstract{
Reconstructing large-scale dynamic driving scenes remains challenging due to the coexistence of static environments with extreme depth variation and diverse dynamic actors exhibiting complex motions. Existing Gaussian Splatting based methods have primarily focused on limited-scale or object-centric settings, and their applicability to large-scale dynamic driving scenes remains underexplored, particularly in the presence of extreme scale variation and non-rigid motions. In this work, we propose DriveSplat, a unified neural Gaussian framework for reconstructing dynamic driving scenes within a unified Gaussian-based representation. For static backgrounds, we introduce a scene-aware learnable level-of-detail (LOD) modeling strategy that explicitly accounts for near, intermediate, and far depth ranges in driving environments, enabling adaptive multi-scale Gaussian allocation. For dynamic actors, we use an object-centric formulation with neural Gaussian primitives, modeling motion through a global rigid transformation and handling non-rigid dynamics via a two-stage deformation that first adjusts anchors and subsequently updates the Gaussians. To further regularize the optimization, we incorporate dense depth and surface normal priors from pre-trained models as auxiliary supervision. Extensive experiments on the Waymo and KITTI benchmarks demonstrate that DriveSplat achieves state-of-the-art performance in novel-view synthesis while producing temporally stable and geometrically consistent reconstructions of dynamic driving scenes.
Project page: \url{https://physwm.github.io/drivesplat}.
}


\maketitle

\begin{figure*}[t]
    \centering
    \includegraphics[width=\linewidth]{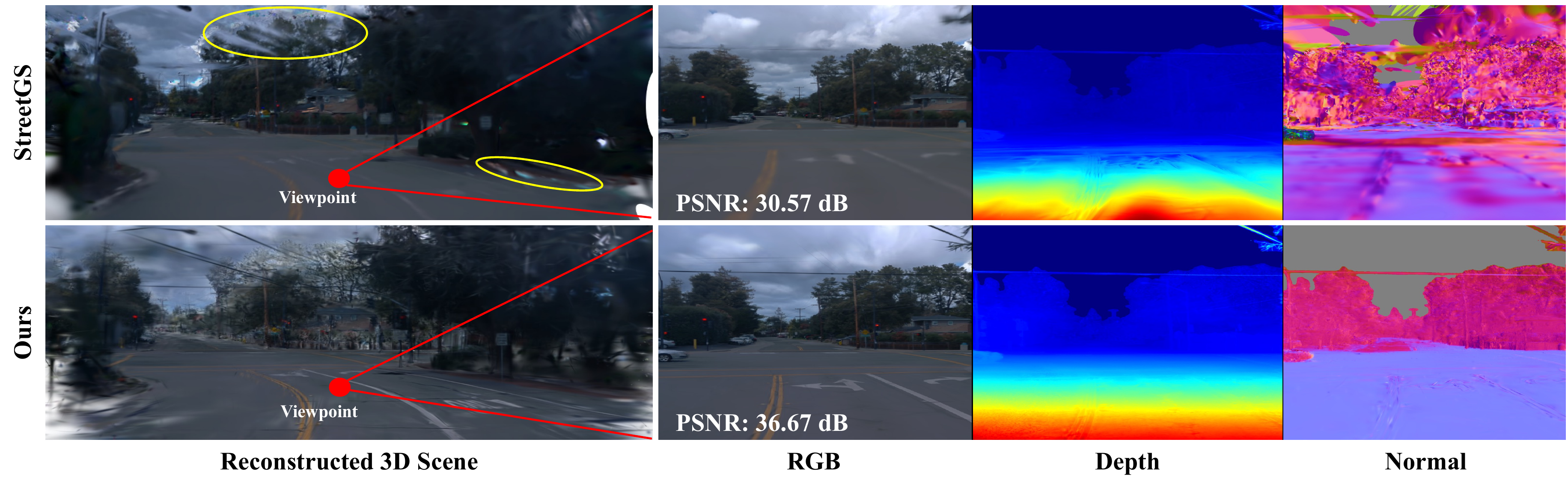}
    \caption{%
        \textbf{Comparison with StreetGS}. 
        StreetGS generates an excessive number of redundant Gaussians (yellow circles) in the reconstructed 3D scene. 
        The right panel presents the rendered image, depth and normal map from a novel view (red dots).
    }
    \label{fig:teaser}
\end{figure*}

\section{Introduction}
\label{sec:intro}

High-fidelity 3D simulation of autonomous driving scenarios plays a crucial role in closed-loop testing and validation of autonomous driving systems. By enabling the flexible construction of complex traffic environments, such simulations also provide valuable training data for perception and decision-making models. Compared to traditional solutions such as oblique photography or manually designed simulators, 3D reconstruction and novel view synthesis (NVS) methods offer a more scalable and realistic alternative by recovering 3D scenes directly from multi-view image observations~\cite{yang2023unisim, zhang2024lp, han2018dense}.

Recent advances in neural scene representations have significantly accelerated progress in this direction~\cite{cho2025dogrecon}. Neural Radiance Fields (NeRF)~\cite{mildenhall2021nerf, barron2021mip} introduce an implicit volumetric representation that achieves high-fidelity view synthesis through differentiable ray sampling~\cite{li2025pseudo, xu2025continuous}. Building upon this paradigm, 3D Gaussian Splatting (3D-GS)~\cite{kerbl20233d} explicitly represents scenes using anisotropic Gaussian primitives, substantially improving rendering efficiency while maintaining competitive visual quality~\cite{zhan2025rdg}. While these methods~\cite{mildenhall2021nerf, kerbl20233d} demonstrate impressive performance in object-centric and small-scale indoor environments, their scalability to large outdoor scenes remains limited. Subsequent works~\cite{yu2024mip, lu2024scaffold, ren2024octree, zhang2024gaussian} improve robustness to viewpoint variations and introduce hierarchical or neural Gaussian representations to better balance reconstruction quality and computational efficiency.

Despite this progress, 3D reconstruction in autonomous driving scenarios remains particularly challenging~\cite{kerbl2024hierarchical, li2024vdg, chen2023periodic, huang2024textit, yan2024street, zhou2024drivinggaussian}. Driving scenes are characterized by large spatial extents, strong scale variation, and the presence of dynamic actors such as vehicles, pedestrians, and cyclists. To address these challenges, recent methods such as StreetGaussian~\cite{yan2024street} and DrivingGaussian~\cite{zhou2024drivinggaussian} adopt dynamic–static decoupling strategies, reconstructing static backgrounds and dynamic foreground actors separately. Follow-up studies further explore non-rigid actor modeling~\cite{chen2024omnire}, trajectory refinement~\cite{ma2025b}, and motion-aware decoupling using optical flow or semantic cues~\cite{sun2025splatflow, xu2025ad}. However, existing representations are not explicitly designed to handle the scale, viewpoint, and geometric characteristics of driving scenes, which manifests as limited robustness under novel viewpoints and suboptimal geometric consistency.

A key limitation of current Gaussian-based driving scene reconstruction methods lies in their treatment of background geometry in large-scale, multi-depth environments. Driving scenes exhibit significant scale variation, where near-range structures demand fine-grained modeling while distant regions primarily require coherent global geometry. However, existing approaches typically rely on single-scale Gaussian densification strategies that improve rendering quality at training viewpoints but tend to introduce redundant primitives in distant regions, degrading visual clarity and geometric consistency under novel viewpoints. This issue is further exacerbated in dynamic driving scenarios, where background reconstruction must remain stable while accommodating the presence of moving and non-rigid actors. Moreover, depth supervision from LiDAR, commonly used in driving datasets, provides sparse and uneven constraints~\cite{chakravarthy2025lidar} that are insufficient for supervising distant structures such as tall buildings and large-scale facades. Surface geometry, particularly normal consistency across scales and viewpoints, is also largely underexplored in existing methods, leading to reconstructions that appear visually plausible yet lack geometric fidelity (see Fig.~\ref{fig:teaser}).

To address these challenges, we propose \textbf{DriveSplat}, a unified neural Gaussian representation framework for robust reconstruction of dynamic driving scenes. Our approach models static backgrounds and dynamic scene elements within a single optimization framework, avoiding ad-hoc design choices tailored to individual components. For background reconstruction, we introduce a scene-aware learnable level-of-detail (LOD) modeling strategy that explicitly accounts for the characteristic near-, mid-, and far-range structure of driving environments. By adaptively allocating multi-scale Gaussian representations, the proposed method enhances the reconstruction of fine-grained geometry in close-range regions while preserving global consistency. Within the same framework, we further extend the representation to dynamic actors. Instead of directly entangling motion with per-Gaussian parameters, we model non-rigid dynamics via an anchor-level deformation mechanism that captures temporal motion through a continuous deformation field and propagates it coherently to neural Gaussian attributes. This design enables stable reconstruction of non-rigid actors while remaining fully compatible with the unified background representation. To further improve geometric accuracy, we incorporate geometry-aware regularization using dense depth and surface normal priors predicted by pretrained monocular models, which effectively enhances both rendering quality and surface consistency. Extensive experiments on the Waymo and KITTI benchmarks demonstrate that our method achieves state-of-the-art performance in novel view synthesis for large-scale driving scenes.

In summary, our main contributions are as follows:
\begin{itemize}
    \item We propose a unified representation for large-scale driving scenes that models both static backgrounds and dynamic actors using neural Gaussians, enabling consistent scene reconstruction under a unified Gaussian-based representation.
    
    \item For background reconstruction, we introduce a scene-aware learnable level-of-detail (LOD) modeling strategy tailored to driving environments, which explicitly accounts for near-, mid-, and far-range scene structures and adaptively allocates multi-scale Gaussian representations to enhance close-range geometric fidelity.
    
    \item For dynamic actors, we design an anchor-level non-rigid deformation mechanism that models temporal motion through a continuous deformation field and propagates it coherently to neural Gaussian parameters, allowing stable reconstruction of non-rigid dynamics within the same representation framework.
    
    \item We further incorporate geometry-aware regularization using dense depth and surface normal priors to improve geometric consistency. Extensive experiments on the Waymo and KITTI benchmarks demonstrate that our method achieves state-of-the-art performance in novel view synthesis for driving scenes.
\end{itemize}

\section{Related Works}
\label{sec:relat}

\subsection{Large-scale Scene Reconstruction}

The naive NeRF \cite{mildenhall2021nerf} struggles with large-scale scenes due to blurry close-ups and jagged distant edges. Improvements include Mip-NeRF \cite{barron2021mip} with multi-scale IPE, NeRF-W \cite{martin2021nerf} for lighting variations, and Block-NeRF \cite{tancik2022block}, which trains local blocks separately.
Recently, 3D-GS-based methods~\cite{kerbl20233d, yu2024mip, liu2025mvsgaussian, zhang2024lp, jiang2025geometry} have achieved remarkable breakthroughs in both reconstruction speed and quality. The initial Gaussian Splatting~\cite{kerbl20233d} is specifically designed for scenes with an object-centered view, and subsequent efforts have extended it to large-scale scenes. 
Neural Gaussian \cite{lu2024scaffold, ren2024octree} incorporates the advantage of Gaussian Splatting and neural fields, and achieves real-time rendering with robust viewpoint invariance.
Hierarchical-GS \cite{kerbl2024hierarchical} introduces a hierarchical structure for driving scenes to optimize the effect of real-time reconstruction and combines the blocking strategy to select different levels.
For the urban scene reconstruction, some methods ~\cite{lin2024vastgaussian, liu2025citygaussian, ren2024octree, peng2024bags} propose to divide point clouds into cells, and introduce the Level-of-Details to optimize reconstruction efficiency and detail performance. 
Above methods neglect dynamic object optimization, while our approach improves reconstruction by decoupling dynamic and static components.

\begin{figure*}[t]
    \centering
    \includegraphics[width=1\linewidth]{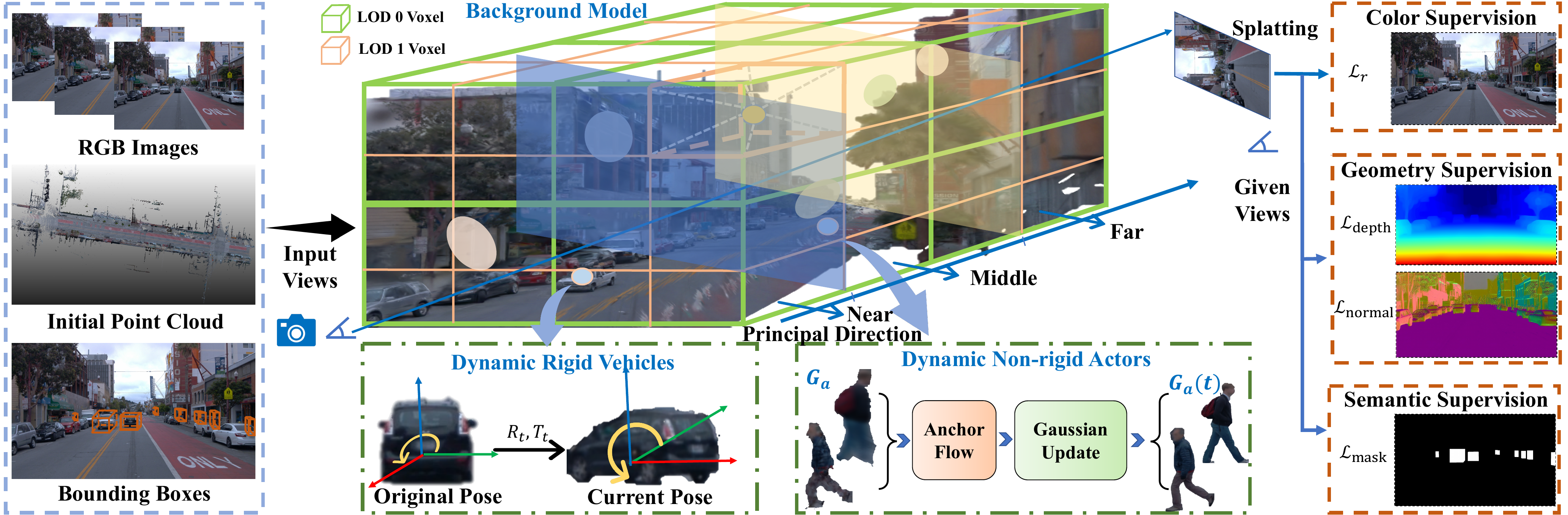}
    \caption{\textbf{Overall pipeline of DriveSplat}. A dynamic-static decoupling paradigm is adopted, where neural Gaussian representations with partitioned voxel structures are applied for background reconstruction, while a deformation field network models the temporal dynamics of each non-rigid actor. Depth maps and normal priors are incorporated to enhance geometric accuracy.}
    \label{fig:pipeline}
\end{figure*}

\subsection{Dynamic Scene Reconstruction}
Traditional reconstruction methods \cite{mildenhall2021nerf, kerbl20233d, yu2024mip} primarily focus on static scenes and are unable to represent dynamic scenes or objects with temporal variations, leading to issues such as motion blur.
The NeRF~\cite{mildenhall2021nerf} leverages MLPs for implicit modeling of static environments. This concept has been extended to animate scenarios through the integration of deformation fields~\cite{guo2023forward, park2021hypernerf, pumarola2021d}. Alternatively, certain strategies~\cite{park2023temporal} conceptualize animate scenes as 4D radiance landscapes, albeit at the cost of significant computational resources attributed to ray-point sampling and volumetric rendering. To mitigate these issues, acceleration techniques~\cite{li2023dynibar, lin2022efficient} have been devised for the depiction of dynamic environments. 
Some methods include the use of geometry priors~\cite{lombardi2021mixture}, the projection of MLP-derived mappings~\cite{peng2023representing}, or the implementation of grid/plane-oriented architectures~\cite{cao2023hexplane} to elevate both the speed and efficacy. 
Several works adapt 3D Gaussians for dynamic scenes \cite{sun2025splatflow, xu2025ad, ma2025b}. Luiten et al.\cite{luiten2023dynamic} train frame-by-frame for multi-view scenes, while Yang et al.\cite{yang2024deformable} use a deformation field to represent the temporal changes of objects. 4D-GS \cite{wu20234d} propose to use multi-resolution hex-planes\cite{cao2023hexplane} to encode deformed motion. 
We learn from the above 4D reconstruction methods and adopt a deformation field to model the temporal evolution of neural Gaussians for non-rigid actors.

\subsection{Geometry Optimization in 3D Reconstruction}
Depth and normal supervision enhance scene reconstruction by improving geometric accuracy and surface orientation, enabling high-fidelity capture of complex scenes~\cite{zhu2023fsgs, jiang2023gaussianshader}.
Several methods \cite{wei2021nerfingmvs, liu2025mvsgaussian} propose to integrate depth priors to guide the reconstruction process. 
The following works \cite{roessle2022dense, wang2023sparsenerf} propose embedding depth supervision into the NeRF framework to boost training efficiency and reduce multi-view input dependency. 
MVSGaussian~\cite{liu2025mvsgaussian} combines MVS with Gaussian Splatting to improve reconstruction in sparse-view settings.
\textcolor{black}{DN-Splatter~\cite{turkulainen2024dn} presents an innovative approach by utilizing depth-normal fusion to enhance point cloud precision in complex environments, while 2D-GS~\cite{huang20242d} leverages 2D depth maps to refine the Gaussian Splatting technique for more efficient reconstruction in real-time applications.} In driving scenarios, GaussianPro~\cite{cheng2024gaussianpro} introduces a progressive propagation strategy that focuses on optimizing geometric properties. And Desire-GS \cite{peng2025desire} proposes the combination of geometric priors for enhanced supervision, but faces problems of very slow training speeds.
Drawing on the above methods, we utilize depth and normal priors to guide neural Gaussian reconstruction, enhancing geometric quality while maintaining reconstruction efficiency.

\section{Methology}
\label{method}

As shown in Fig. \ref{fig:pipeline}, the inputs to DriveSplat include RGB images, an initialized 3D point cloud, and bounding boxes of dynamic actors provided by the dataset~\cite{sun2020scalability, geiger2012we}. Depth and normal priors predicted by pre-trained models~\cite{yang2024depthv2, eftekhar2021omnidata} are utilized during the supervised optimization stage.

\subsection{Point Cloud Initialization}
\label{subsec:point}
DriveSplat supports multiple types of point cloud initialization, including SfM, LiDAR, and dense DUSt3R \cite{wang2024dust3r} input.
Our model separates dynamic actors by leveraging tracked bounding boxes to estimate their pose parameters ($\mathbf{q}_{\text{obj}}$, $\mathbf{t}_{\text{obj}}$), which are used to compute a transformation matrix $\mathbf{T}_{\text{obj}}$. When LiDAR data is available, each frame's point cloud $\mathbf{P}_{\text{frame}}$ is then transformed into the actor's local coordinate system for consistent modeling. Within the local coordinate system, an axis-aligned bounding box $\mathcal{B}$ is constructed to identify and filter the points $\mathbf{P}_{\text{obj}}$ enclosed within it. \textcolor{black}{This process is applied to all tracked actors, generating dynamic actor point cloud masks $\mathbf{M}_{\text{obj}}$}.
\begin{equation}
    \mathbf{P}_{\text{obj}} = \bigcup_{i} \left\{ \mathbf{P}_{\text{frame}}^{(i)} \mathbf{T}_{\text{obj}}^{-1} \mid \mathbf{M}_{\text{obj}}^{(i)} = 1 \right\}.
    \label{eq:p_obj}
\end{equation}

The remaining unmasked points are classified as static points, defined as:
\begin{equation}
    \mathbf{P}_{\text{s}} = \{ \mathbf{p} \mid \mathbf{p} \in \mathbf{P}_{\text{frame}}, \mathbf{M}_{\text{obj}}(\mathbf{p}) = 0 \}.
    \label{eq:p_bkgd}
\end{equation}

In the absence of LiDAR data, our method initializes the point cloud of dynamic actors by randomly initializing points within the bounding boxes in the relative coordinate system. 
For static background points, when LiDAR is not used, we directly use COLMAP's point cloud for initialization, as it contains only static points.

\subsection{Large-scale Driving Background Representation}
\label{subsec:background}
We propose a scene-aware background reconstruction framework for large-scale driving environments based on neural Gaussian representations.
The framework explicitly accounts for the geometric characteristics of driving scenes, including large spatial extent, strong depth variation, and continuous camera motion.

\begin{figure}[t]
    \centering
    \includegraphics[width=1.0\linewidth]{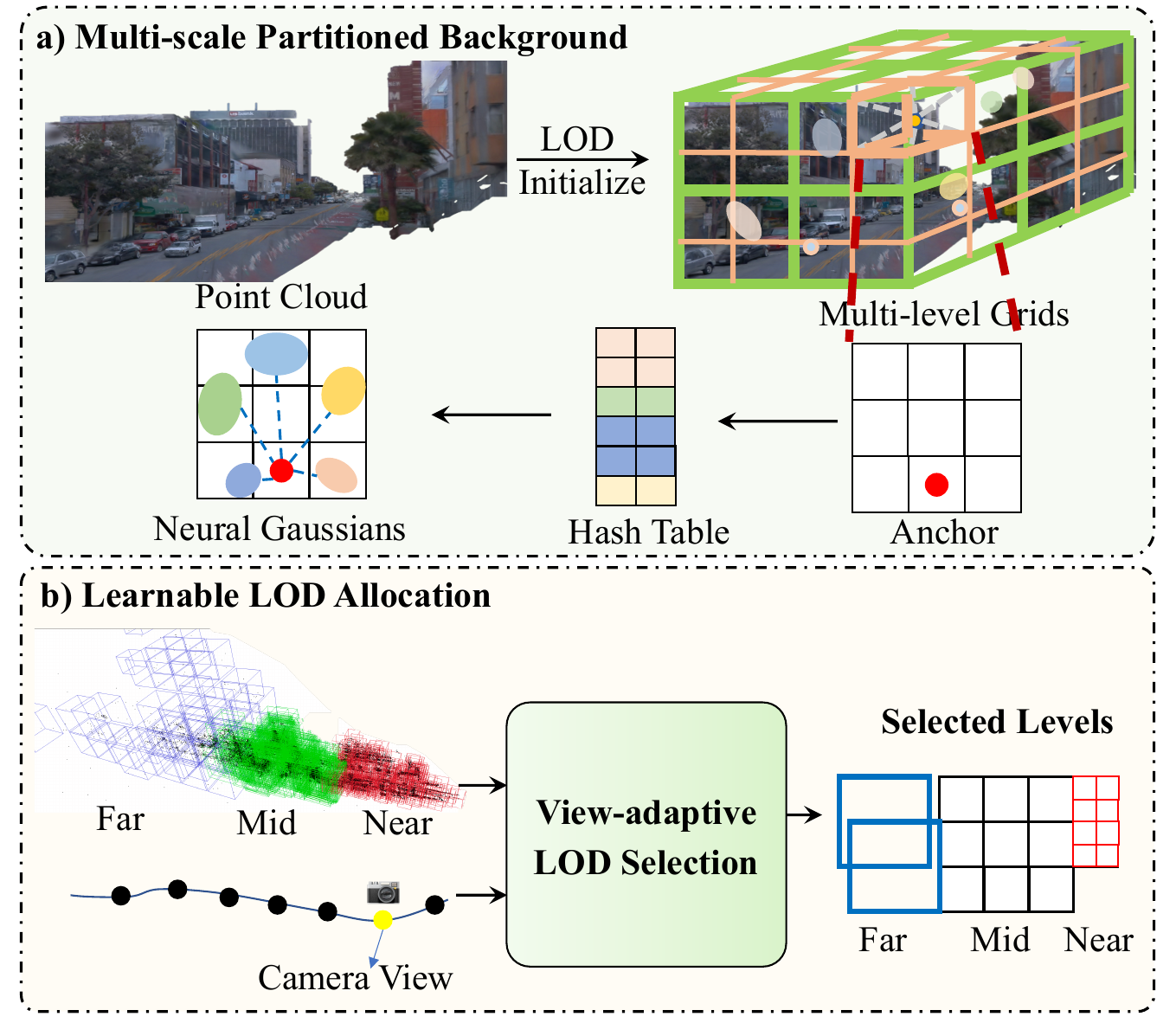}
    \caption{\textbf{Overview of the proposed multi-scale background representation and view-adaptive LOD allocation}.
(a) The static background is modeled using a multi-scale Gaussian representation.
(b) Geometry-guided near/mid/far regions provide structural priors, while the effective LOD of each anchor is dynamically selected based on the current camera viewpoint.}
    \label{fig:static}
\end{figure}

Our approach integrates a unified multi-scale Gaussian feature representation, geometry-guided scene partitioning, and a view-adaptive LOD allocation strategy.
Together, these components enable robust background reconstruction under significant viewpoint changes.
An overview of the proposed framework is illustrated in Fig.~\ref{fig:static}.

\paragraph{Scene-aware multi-scale Gaussian representation.}
\label{subsubsec:hash_encoding}\
To effectively model the large-scale variation inherent in driving scenes, we adopt a multi-scale Gaussian representation in which each spatial anchor is associated with level-specific latent features.
Rather than relying on explicit hierarchical spatial data structures~\cite{ren2024octree}, we parameterize anchor features across multiple scales using a unified and learnable representation that remains flexible to scene complexity.

Specifically, we maintain a set of $L$ resolution levels, each corresponding to a distinct spatial scale.
At each level $l \in \{0, \dots, L-1\}$, anchor features are parameterized by a learnable table $T^{(l)} \in \mathbb{R}^{H \times F}$, where $H$ denotes the table capacity and $F$ the feature dimensionality.
The spatial resolution associated with level $l$ is defined as
\begin{equation}
    r^{(l)} = \lfloor r_{\text{base}} \cdot b^{l/(L-1)} \rfloor,
    \label{eq:resolution}
\end{equation}
where $r_{\text{base}}$ and $r_{\text{finest}}$ denote the minimum and maximum spatial resolutions considered in the multi-scale representation, and $b = r_{\text{finest}} / r_{\text{base}}$ defines the geometric scaling factor between consecutive resolution levels.
This formulation enables the representation to accommodate spatial structures of varying scale within a single parameterization, which is essential for modeling both fine-grained geometry and large-scale layout in driving environments.

To associate spatial anchors with the corresponding level-specific feature parameters, we define a mapping from continuous 3D positions to discrete feature indices at each resolution level.
Given a 3D anchor position $\mathbf{x} \in \mathbb{R}^3$, we normalize it within the scene bounding box and discretize it into integer grid coordinates at level $l$ as:
\begin{equation}
    \mathbf{g}^{(l)} = \left\lfloor \frac{\mathbf{x} - \mathbf{x}_{\min}}{\mathbf{x}_{\max} - \mathbf{x}_{\min}} \cdot r^{(l)} \right\rfloor.
    \label{eq:quantize}
\end{equation}
The resulting grid coordinates are then mapped to feature indices using a spatial hashing function
\begin{equation}
    h(\mathbf{g}) = (g_x \cdot \pi_1 \oplus g_y \cdot \pi_2 \oplus g_z \cdot \pi_3) \bmod H,
    \label{eq:hash}
\end{equation}
where $\pi_1, \pi_2, \pi_3$ are fixed large prime numbers and $\oplus$ denotes the bitwise XOR operator.

Each anchor is associated with a single level-of-detail assignment $l$, which is optimized by a scene-aware LOD allocation strategy discussed later.
Given an anchor position $\mathbf{a}$ and its assigned level $l$, the corresponding feature vector is obtained as:
\begin{equation}
    \mathbf{f}_{\text{hash}}(\mathbf{a}, l) = T^{(l)}\big[h(\mathbf{g}^{(l)}(\mathbf{a}))\big].
    \label{eq:feature_query}
\end{equation}
By enforcing level-specific feature access rather than aggregating features across multiple resolutions, each anchor is represented at a single, well-defined spatial scale.
This design reduces implicit parameter coupling across scales and leads to a more stable and interpretable multi-scale Gaussian representation, which is particularly suitable for large-scale driving scenes.

\begin{figure}[t]
    \centering
    \includegraphics[width=1.0\linewidth]{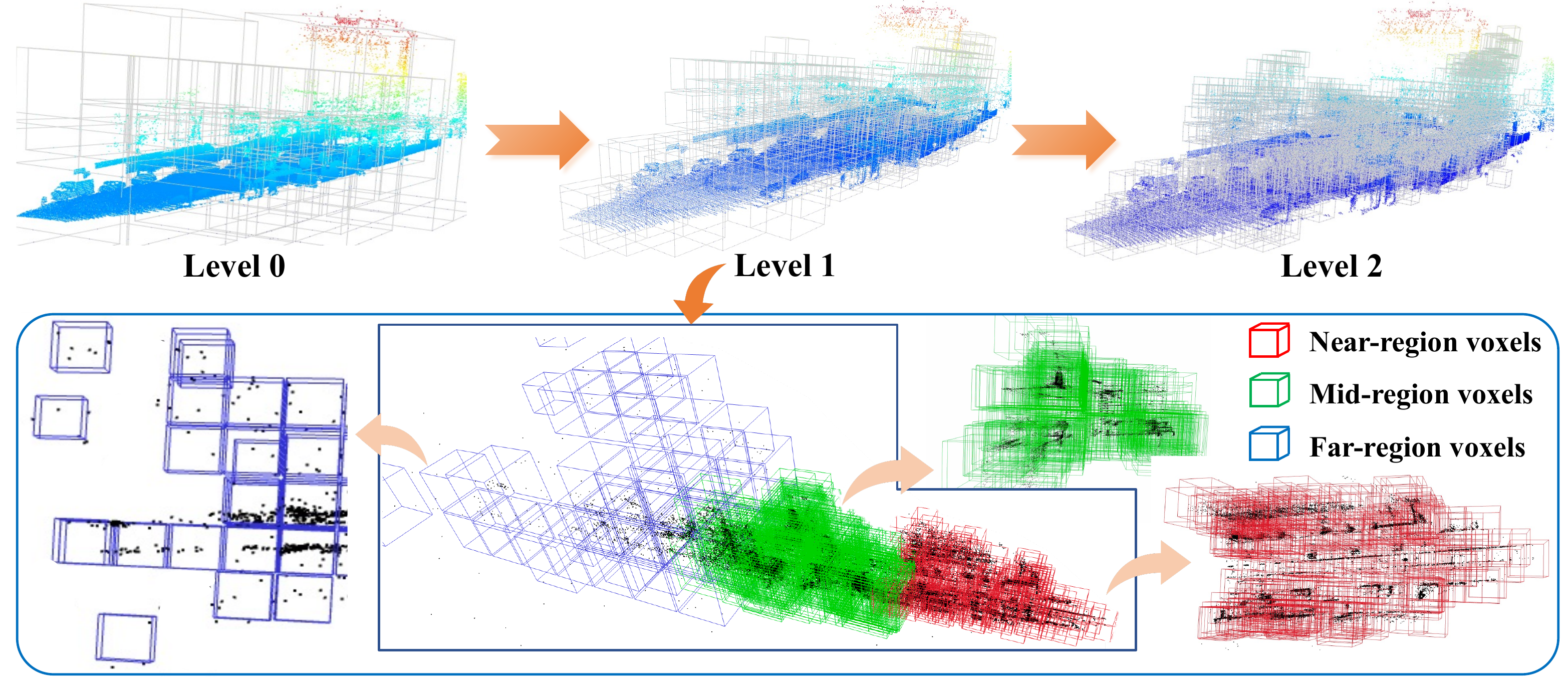}
    \caption{\textbf{Visualization of voxel representation at different levels}. As the level increases, the \textcolor{black}{voxel resolution} gradually improves. The corresponding partitioned neural Gaussians are shown in the bottom row.}
    \label{fig:voxel}
\end{figure}

\paragraph{Geometry-guided scene partitioning for driving environments.}
\label{subsubsec:partitioning}
Driving scenes span a wide range of depths, with scene elements ranging from regions close to the ego vehicle to distant urban structures.
Such characteristics are not well captured by uniform spatial discretization strategies, which treat all spatial regions equivalently and ignore the depth-dependent structure inherent in driving environments.
To better align the representation with scene geometry, we introduce a geometry-guided partitioning strategy that decomposes the background into three depth-ordered regions, referred to as the near, mid, and far regions.

Given a set of 3D points $\mathcal{P} = \{\mathbf{p}_i\}_{i=1}^N$ sampled from the background point cloud $\mathbf{P}_{\text{s}}$, we first estimate the dominant geometric direction of the scene via principal component analysis (PCA):
\begin{equation}
    \mathbf{d}_{\text{main}} = \arg\max_{\|\mathbf{v}\|=1} \mathrm{Var}\big(\mathbf{v}^\top (\mathcal{P} - \bar{\mathcal{P}})\big),
    \label{eq:pca}
\end{equation}
where $\bar{\mathcal{P}}$ denotes the centroid of the point cloud.
In typical driving scenarios, the resulting principal direction $\mathbf{d}_{\text{main}}$ aligns closely with the forward viewing direction of the ego vehicle, providing a meaningful reference for depth-based scene decomposition.

We then project each point $\mathbf{p}_i$ onto the principal direction to obtain a scalar depth coordinate $z_i$,
and partition the depth distribution into three regions using quantile-based thresholds:
\begin{equation}
\begin{aligned}
    \mathcal{R}_{\text{near}} &= \{\mathbf{p}_i : z_i \leq \tau_1\}, \\
    \mathcal{R}_{\text{mid}}  &= \{\mathbf{p}_i : \tau_1 < z_i \leq \tau_2\}, \\
    \mathcal{R}_{\text{far}}  &= \{\mathbf{p}_i : z_i > \tau_2\},
\end{aligned}
\label{eq:region}
\end{equation}
where $\tau_1$ and $\tau_2$ correspond to the $q_1$-th and $q_2$-th percentiles of the depth distribution, respectively.
Unless otherwise specified, we set $q_1 = 0.33$ and $q_2 = 0.67$ in our experiments.
An example of the resulting partitioning is visualized in Fig.~\ref{fig:voxel}.

Based on this partitioning, we assign region-specific spatial resolutions to better match the geometric characteristics of different depth ranges.
Specifically, the effective voxel size for region $r \in \{\text{near}, \text{mid}, \text{far}\}$ is defined as:
\begin{equation}
    s_r = s_{\text{base}} / \alpha_r,
    \label{eq:voxel_size}
\end{equation}
where $s_{\text{base}}$ denotes a reference spatial scale and $\alpha_{\text{near}} > \alpha_{\text{mid}} > \alpha_{\text{far}}$ are region-dependent scaling factors.
This design enables finer geometric representation in regions closer to the camera while maintaining coherent structure in more distant areas, and serves as a geometry-aware foundation for subsequent level-of-detail allocation.

\paragraph{Learnable Level-of-Detail allocation.}
\label{subsubsec:lod_allocation}
Assigning LOD based solely on static heuristics or initialization-time region labels is insufficient for large-scale driving scenes, as camera motion continuously alters the visual relevance of different spatial regions.
Although the background scene is first decomposed into near-, mid-, and far-range regions using geometry-guided partitioning, this coarse structural prior alone does not fully determine the appropriate representation granularity.
Anchors initially assigned to distant regions may later become visually prominent as the camera approaches them, motivating a view-adaptive LOD allocation strategy that dynamically selects the effective representation level of each anchor.

To account for this effect, we dynamically select anchor visibility and LOD levels at each rendering frame based on the current camera viewpoint.
For an anchor $i$, we compute a target LOD level as
\begin{equation}
    L_i^{\text{target}} = \left\lfloor \log_s \frac{d_{\max}}{d_i} \right\rfloor,
\end{equation}
where $d_i$ denotes the Euclidean distance between anchor $i$ and the camera center of current view, $d_{\max}$ is a scene-dependent normalization constant, and $s$ controls the LOD scaling factor.
An anchor is rendered if and only if its assigned resolution level $L_i$ satisfies $L_i \leq L_i^{\text{target}}$.
This formulation ensures that anchors are selected according to their instantaneous visual relevance rather than fixed depth labels.

The geometry-guided scene partitioning introduced in Sec.~\ref{subsubsec:partitioning} provides a coarse structural prior by decomposing the background into distinct regions.
While this partitioning offers a reasonable initialization, fixed region boundaries are insufficient to capture the continuously changing visual relevance induced by camera motion in driving scenes.
We therefore lift the partition boundaries to learnable parameters and refine them jointly with the LOD allocation during training.

Concretely, the region boundaries $\tau_1$ and $\tau_2$ are treated as continuous learnable variables, initialized using the quantile-based geometry partitioning.
Given the projected depth coordinate $z_i$ of anchor $i$, its soft region assignment is defined as
\begin{equation}
    w_{i,r} = \sigma\!\left( \frac{z_i - \tau_{r-1}}{\ell_r} \right) - 
              \sigma\!\left( \frac{z_i - \tau_r}{\ell_r} \right),
    \label{eq:soft_region}
\end{equation}
where $r \in \{\text{near}, \text{mid}, \text{far}\}$, $\sigma(\cdot)$ denotes the sigmoid function, $\ell_r$ controls the transition sharpness between adjacent regions, and we define $\tau_0 = -\infty$ and $\tau_3 = +\infty$.
This formulation yields a differentiable, region-aware weighting that allows anchors to smoothly adjust their effective LOD preferences as the region boundaries evolve.

The parameters $\{\tau_1, \tau_2, \ell_r\}$ are optimized jointly with the neural Gaussian representation via gradients propagated from the reconstruction objective, which consists of photometric and geometry-aware losses.
Through this reconstruction-driven supervision, the initially geometry-defined partitions are gradually refined to better align with view-dependent visibility and scene content, enabling a data-driven and adaptive LOD allocation without relying on manually tuned heuristics.

To ensure stable optimization, we enforce the ordering constraint $\tau_1 < \tau_2$ by parameterizing them as cumulative offsets.
Unless otherwise specified, all learnable partition parameters are initialized to match the geometry-guided partitioning described in Sec.~\ref{subsubsec:partitioning}.

\subsection{Dynamic Actor Modeling}

\begin{figure*}[t]
    \centering
    \includegraphics[width=0.9\linewidth]{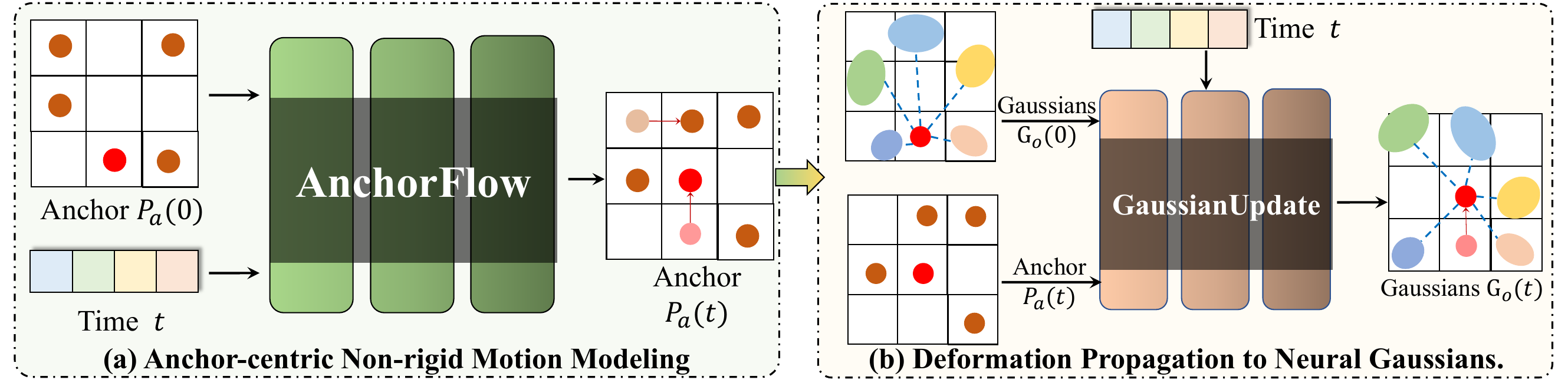}
    \caption{\textbf{Pipeline for dynamic non-rigid actor modeling}. Our two-stage pipeline first estimates anchor-level motion and then updates the corresponding neural Gaussian parameters.}
    \label{fig:pipe_flow}
\end{figure*}

Dynamic actors in driving scenes include both rigid objects, such as vehicles, and non-rigid actors, such as pedestrians, which exhibit complex combinations of global motion and local deformation.
To model these dynamics in a unified manner, we adopt an object-centric formulation that decomposes actor motion into a shared global rigid transformation and an optional non-rigid deformation component.

\paragraph{Global actor motion.}
For each actor, we first model its global motion as a time-varying rigid transformation that aligns the actor from a canonical coordinate system to the world frame.
This global transformation is shared by both rigid and non-rigid actors and captures their overall translation and rotation over time.

The position \(\boldsymbol{\mu}_o\) and rotation \(\mathbf{R}_o\) for an actor are defined within the actor's local coordinate system. To align with the background, these variables must be transformed into the world coordinate system, which requires applying the actor's tracked poses. Specifically, these tracked poses are represented by a set of rotation matrices \(\{\mathbf{R}_t\}^{N_t}_{t=1}\) and translation vectors \(\{\mathbf{T}_t\}^{N_t}_{t=1}\), where \(N_t\) denotes the total number of frames. This transformation is expressed as
\begin{equation}
    \begin{aligned}
        \boldsymbol{\mu} &= \mathbf{R}_t\boldsymbol{\mu}_o + \mathbf{T}_t,\\
        \mathbf{R} &= \mathbf{R}_o\mathbf{R}_t^T,
    \end{aligned}
\end{equation}
where \(\boldsymbol{\mu}\) and \(\mathbf{R}\) denote the position and rotation of the dynamic actor's Gaussians $\mathbf{G}_a$ within the world coordinate system.

At each time step \(t\), these canonical Gaussian parameters are mapped to the world coordinate system via the rigid transformation, ensuring consistent pose alignment throughout the sequence.

\paragraph{Anchor-centric non-rigid motion modeling.}
While rigid actors are fully described by the global transformation alone, non-rigid actors require additional modeling to capture local deformations.
Rather than directly modeling dense per-Gaussian deformations in 4D space, we observe that, in anchor-based neural Gaussian representations, the spatiotemporal evolution of Gaussians is primarily governed by the motion of their associated anchors.

Motivated by this observation, we introduce \textbf{AnchorFlow}, an N-layer MLP that learns a continuous flow field over a sparse set of anchors in canonical space.
Conditioned on anchor representations and an encoded time variable \(t\), AnchorFlow predicts the updated anchor positions \(P_a(t)\), providing a compact and differentiable parameterization of non-rigid motion over time, as illustrated in Fig.~\ref{fig:pipe_flow}(a).

This anchor-centric formulation introduces a strong inductive bias that enforces coherent motion among neural Gaussians linked to the same anchor, while substantially reducing the number of learnable parameters compared to dense 4D Gaussian deformation models.

\paragraph{Deformation propagation to neural Gaussians.}
Given the updated anchor states at time \(t\), we propagate anchor motion to the associated neural Gaussians through a multi-head update scheme.

Specifically, a geometry update head synchronizes the spatial attributes of each Gaussian with its corresponding anchor, updating spatial attributes based on the anchor displacement and temporal encoding, as shown in Fig.~\ref{fig:pipe_flow}(b). This anchor-driven synchronization ensures coherent geometric deformation among Gaussians linked to the same anchor.

Following the geometric update, the appearance heads independently update opacity $\alpha$ and color $c$ for each neural Gaussian, \textcolor{black}{producing time-dependent appearance parameters \(\alpha(t)\) and \(c(t)\)}. By decoupling geometry propagation from appearance evolution, our design enables stable optimization while preserving flexible, per-Gaussian appearance dynamics under non-rigid motion.

\paragraph{Compositional scene rendering.}

After applying the rigid transformation and the optional non-rigid deformation, we obtain the updated dynamic actor representation $\mathbf{G}_a(t)$ at timestamp $t$. We then combine the dynamic Gaussians of all actors with the static background Gaussians $\mathbf{G}_b$ to form the complete scene representation:
\begin{equation}
    \mathbf{G}(t) = (\cup_{i=1}^{K} \mathbf{G}_a^{i}(t)) \cup \mathbf{G}_b,
\end{equation}
where $K$ denotes the number of dynamic actors reconstructed in the scene. Finally, given the camera view matrix $M_t$ of the current frame, we render the RGB image $I_t$, depth map $D_t$, and normal map $N_t$ using the splatting operator $\mathcal{S}$:
\begin{equation}
    (I_t, D_t, N_t) = \mathcal{S}(M_t, \mathbf{G}(t)).
\end{equation}

\subsection{Geometry Enhanced Optimization}

\paragraph{Geometry prior estimation.}
We utilize depth and normal priors to guide the geometry optimization process.
To obtain per-pixel referenceable depth values, monocular depth estimation can be employed to predict either absolute depth $D_{\text{m}}$ or relative depth $D_{\text{r}}$. We use the DepthAnything-V2~\cite{yang2024depthv2} model for relative depth estimation and the ZoeDepth~\cite{bhat2023zoedepth} model for absolute depth estimation.
To obtain reliable surface normal priors, we leverage a pre-trained normal estimation model \cite{eftekhar2021omnidata} to generate normal maps $N_{\text{m}}$. These estimated normals provide crucial geometric constraints, contributing to improved surface quality of the reconstructed driving scenario.

\paragraph{Loss functions.}\label{sec:loss}
To achieve a high-quality rendering, we define a comprehensive loss function as:
\begin{equation}
    \mathcal{L} = \mathcal{L}_{\text{r}} + \lambda_{\text{d}} \mathcal{L}_\text{depth} + \lambda_{\text{n}} \mathcal{L}_\text{normal} + \lambda_{\text{m}} \mathcal{L}_\text{mask},
    \label{loss:total}
\end{equation}
where $\mathcal{L}_{\text{r}}$, $\mathcal{L}_\text{depth}$, $\mathcal{L}_\text{normal}$, $\mathcal{L}_\text{mask}$ represent the rendered color loss, depth loss, normal loss, and dynamic actor mask loss, respectively. Here, \(\lambda_{\text{d}}\), \(\lambda_{\text{n}}\), and \(\lambda_{\text{m}}\) are hyperparameters that control the weights of each loss term during optimization. 

The rendering color loss \(\mathcal{L}_{\text{r}}\) 
incorporates an L1 loss \(\mathcal{L}_{\text{L1}}(I, I_\text{gt})\) and a structural similarity index loss \(\mathcal{L}_\text{SSIM}(I, I_\text{gt})\), 
defined as:
\begin{equation}
    \mathcal{L}_{\text{r}} = \mathcal{L}_{\text{L1}}(I, I_\text{gt}) + \lambda  \mathcal{L}_\text{SSIM}(I, I_\text{gt}),
    \label{loss:rgb}
\end{equation}
where \(\lambda\) balances the contribution of the L1 loss and SSIM loss to achieve high-quality rendering results. 

To further enhance geometric accuracy, we include the depth loss \(\mathcal{L}_\text{depth}\), to enforce consistency between our estimated depth map \(D\) and the predicted relative depth map \(D_\text{r}\). This loss is scaled by weight \(\lambda_{d}\) and computed using correlation loss:
\begin{equation}
    \mathcal{L}_\text{depth} = \|\frac{\text{Cov}(D, D_\text{r})}{\sqrt{\text{Var}(D)\text{Var}(D_\text{r})}}\|_1,
    \label{loss:depth}
\end{equation}
where $\text{Cov}(\cdot)$ and $\text{Var}(\cdot)$ denoting the covariance and variance, respectively. Our method also supports supervision using absolute depth \(D_\text{m}\), where the L1 loss is used rather than the correlation loss.

To ensure that the predicted surface normals align with the real surface normal distribution, we incorporate the normal loss \(\mathcal{L}_\text{normal}\). The loss includes L1 loss $\mathcal{L}_{n_{\text{L1}}}$ for error between our estimated normal $N$ and predicted reference normal $N_\text{m}$, and cosine similarity loss $\mathcal{L}_{n_{\text{cos}}}$ for normal direction alignment. The total normal loss is:
\begin{equation}
    \mathcal{L}_\text{normal} = \mathcal{L}_{n_{\text{L1}}}(N, N_\text{m}) + \mathcal{L}_{n_\text{cos}}(N, N_\text{m}).
    \label{loss:normal}
\end{equation}

In addition, the mask loss \(\mathcal{L}_\text{mask}\) employs cross-entropy to compare predicted and ground-truth masks, helping to improve rendering quality of dynamic actors.

\begin{figure*}[t]
    \centering
    \includegraphics[width=1\linewidth]{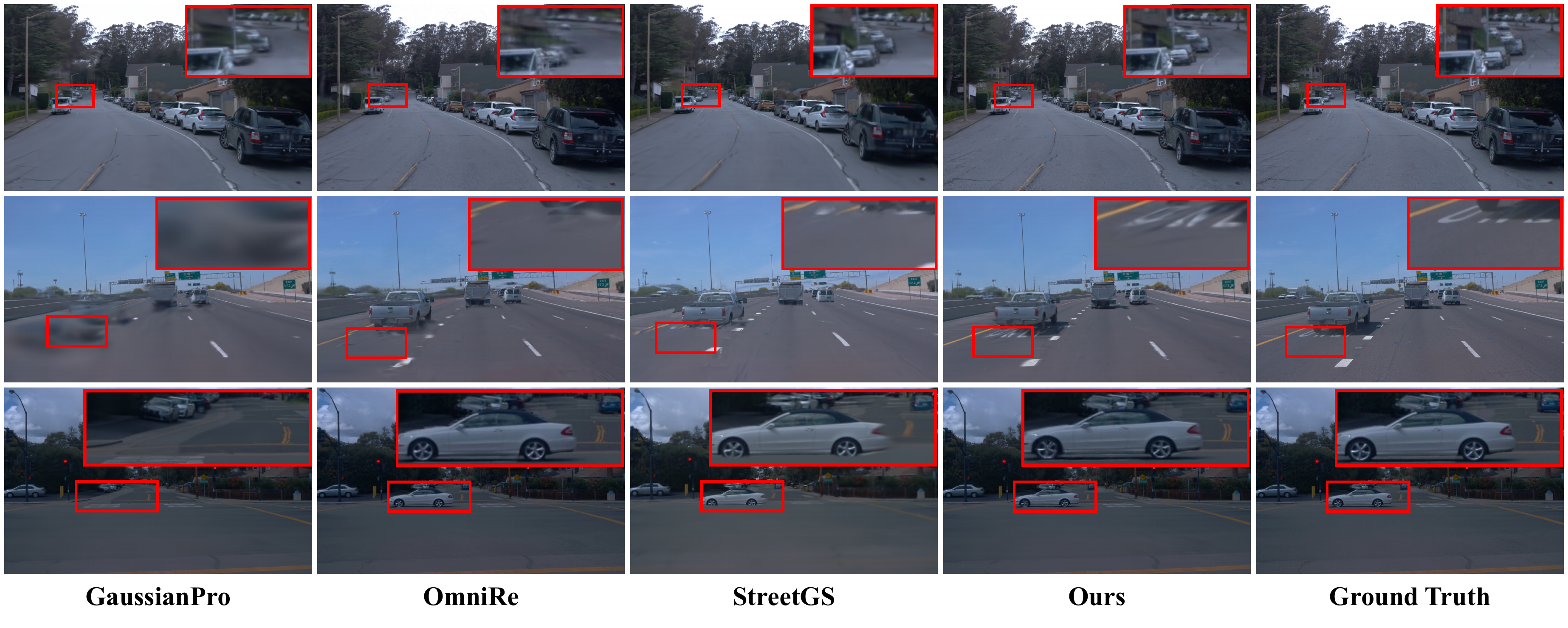}
    \caption{\textbf{Qualitative novel-view synthesis results on the Waymo dataset}. Patches that highlight the visual differences are emphasized with \textcolor{black}{red} boxes and enlarged for clearer visibility. The rendering image resolution is $1066 \times 1600$.}
    \label{fig:com_waymo}
\end{figure*}

\begin{table*}[t]
    \centering
    \caption{\textbf{Quantitative comparison results on Waymo dataset}. S and D indicate methods that model static scenes only and dynamic scenes, respectively. $\mathbf{D_{opt}}$ and $\mathbf{N_{opt}}$ denote depth and normal optimization, respectively. The image resolution is $1066 \times 1600$. LPIPS uniformly adopts the AlexNet. \textbf{Bold}: Best. \underline{Underline}: Second Best.}
    \setlength{\tabcolsep}{6pt}
    \renewcommand{\arraystretch}{1.0} 
    \begin{tabularx}{\linewidth}{@{} lcccccccccc @{}}
    \toprule
    \multirow{2}{*}{\textbf{Model}} & \multirow{2}{*}{\textbf{Type}} & \multirow{2}{*}{$\mathbf{D_{opt}}$} & \multirow{2}{*}{\textbf{$\mathbf{N_{opt}}$}} & \multicolumn{3}{c}{\textbf{Reconstruction}} & \multicolumn{3}{c}{\textbf{NVS}} \\
    \cmidrule(lr){5-7} \cmidrule(lr){8-10}
         & & & & PSNR↑ & SSIM↑ & LPIPS↓ & PSNR↑ & SSIM↑ & LPIPS↓ \\ \midrule
        3D-GS \cite{kerbl20233d} & S & $\times$ & $\times$ & 33.15 & 0.929 & 0.108 & 30.57 & 0.923 & 0.113 \\
        GaussianPro \cite{cheng2024gaussianpro} & S & $\checkmark$ & $\checkmark$ & 32.79 & 0.928 & 0.113 & 31.28 & 0.915 & 0.121 \\
        Octree-GS \cite{ren2024octree} & S & $\times$ & $\times$ & 33.12 & 0.934 & 0.109 & 31.84 & 0.917 & \underline{0.109} \\
        4D-GS \cite{wu20234d} & D & $\times$ & $\times$  & 31.15 & 0.897 & 0.197 & 28.35 & 0.873 & 0.208 \\
        Deform-GS \cite{yang2024deformable} & D & $\times$ & $\times$ & 32.58 & 0.909 & 0.162 & 30.04 & 0.892 & 0.174 \\
        S3Gaussian \cite{huang2024textit} & D & $\checkmark$ & $\times$ & 33.64 & 0.931 & 0.117 & 31.32 & 0.912 & 0.127 \\
        PVG \cite{chen2023periodic} & D & $\checkmark$ & $\times$ & 34.37 & 0.934 & \underline{0.102} & 31.89 & 0.912 & 0.118 \\
        StreetGS \cite{yan2024street} & D & $\times$ & $\times$ & 35.15 & 0.935 & 0.110 & 30.24 & 0.878 & 0.125 \\
        OmniRe \cite{chen2024omnire} & D & $\checkmark$ & $\times$ & 34.57 & \textbf{0.939} & \textcolor{black}{0.112} & 31.19 & 0.897 & 0.126 \\
        Desire-GS \cite{peng2025desire} & D & $\checkmark$ & $\checkmark$ & 34.35 & 0.925 & 0.109 & 32.35 & 0.917 & 0.122 \\
        AD-GS \cite{xu2025ad} & D & $\checkmark$ & $\times$ & \underline{35.26} & 0.936 & 0.105 & \underline{33.08} & \underline{0.920} & 0.112 \\
        \midrule
        \textbf{Ours} & D & $\checkmark$ & $\checkmark$ & \textbf{35.41} & \underline{0.937} & \textbf{0.096} & \textbf{33.83} & \textbf{0.923} & \textbf{0.103} \\ 
        \bottomrule
    \end{tabularx}
    \label{tab:waymo}
\end{table*}

\begin{figure*}[ht]
    \includegraphics[width=1.0\textwidth]{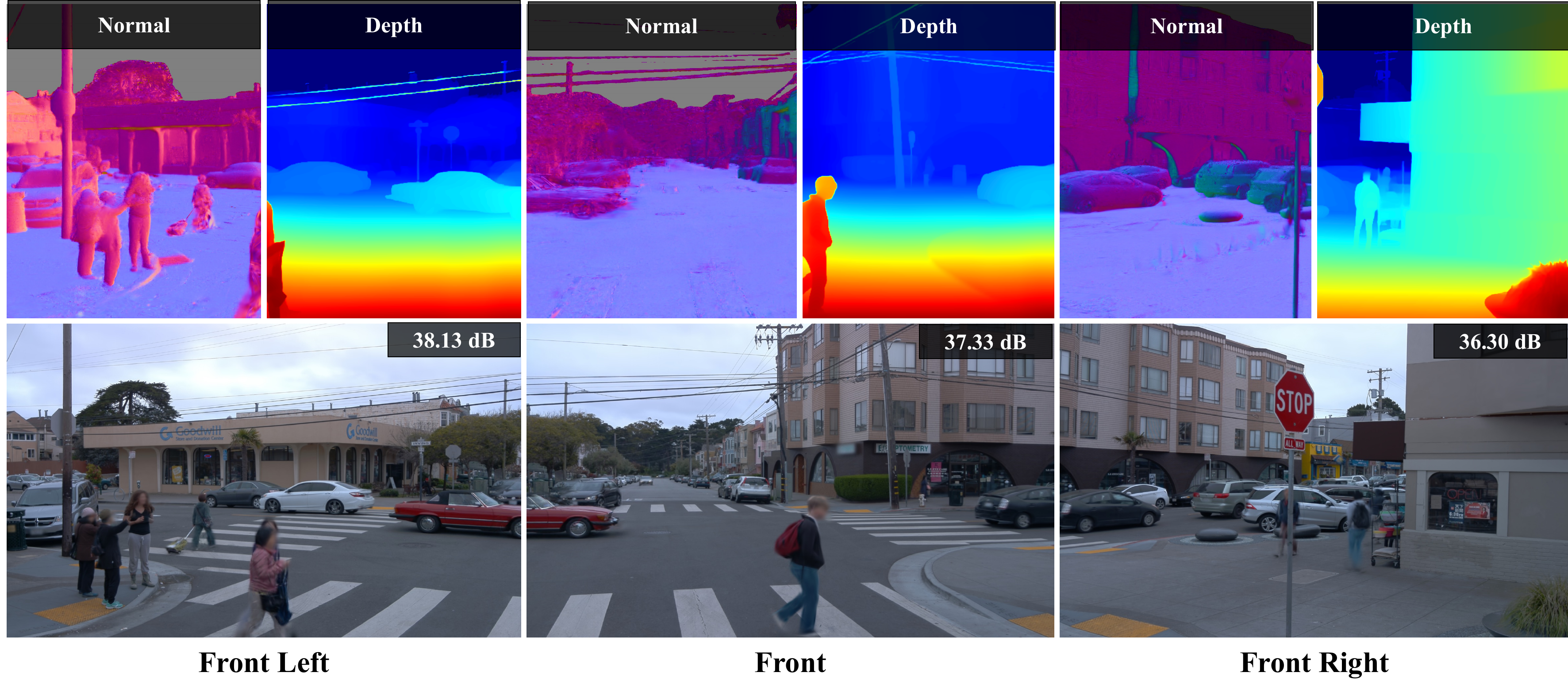}
    \vspace{-0.35cm}
    \caption{\textbf{Front 3-views rendering results of our method on the Waymo dataset}. The above shows the rendered depth and normal maps, and below shows the rendered images. Our method can preserve the intricate details of the scene, and generate accurate depth and normal maps.}
    \label{fig:teaser1}
\end{figure*}

\begin{table*}
    \centering
    \caption{\textbf{Quantative results on KITTI dataset}. The rendering image resolution is $370 \times 1226$.}
    \setlength{\tabcolsep}{11pt}
    \renewcommand{\arraystretch}{1.0} 
    \begin{tabularx}{\linewidth}{@{} lcccccccc @{}}
    \toprule
    \multirow{2}{*}{\textbf{Model}}  & \multirow{2}{*}{\textbf{Type}} & \multicolumn{3}{c}{\textbf{Scene Reconstruction}} & \multicolumn{3}{c}{\textbf{Novel View Synthesis}} \\
    \cmidrule(lr){3-5} \cmidrule(lr){6-8}
         & & PSNR↑ & SSIM↑ & LPIPS↓ & PSNR↑ & SSIM↑ & LPIPS↓ \\ 
         \midrule
        3D-GS~\cite{kerbl20233d} & S & 24.43 & 0.817 & 0.162 & 20.10 & 0.678 & 0.224 \\
        Octree-GS~\cite{ren2024octree} & S & 23.51 & 0.738 & 0.288 & 21.90 & 0.679 & 0.300 \\
        GaussianPro~\cite{cheng2024gaussianpro}& S & 23.51 & 0.788 & 0.181 & 18.47 & 0.607 & 0.277 \\
        4D-GS~\cite{wu20234d}& D  & 24.43 & 0.735 & 0.280 & 18.12 & 0.562 & 0.338 \\
        Deform-GS~\cite{yang2024deformable}& D  & 27.64 & 0.810 & \underline{0.113} & \underline{22.01} & \underline{0.727} & \underline{0.156} \\
        OmniRe~\cite{chen2024omnire}& D & \textbf{28.68} & \underline{0.874} & 0.115 & 20.86 & 0.592 & 0.187 \\
        \midrule
        \textbf{Ours} & D & \underline{28.59} & \textbf{0.895} & \textbf{0.100} & \textbf{24.53} & \textbf{0.767} & \textbf{0.142} \\ 
        \bottomrule
    \end{tabularx}
    \label{tab:kitti}
\end{table*}

\section{Experiments}
\subsection{Datasets}

\noindent \textbf{Waymo}~\cite{sun2020scalability} provides a diverse collection of sensor data covering both urban and suburban environments. We select 12 sequences captured under a variety of conditions, including different weather settings (e.g., foggy and sunny) and traffic scenarios (e.g., urban low-speed roads and highways).
Among them, the representative set of 12 sequences consists of 8 sequences from StreetGS~\cite{yan2024street} and 4 sequences from OmniRe~\cite{chen2024omnire}. This combination is intentionally chosen to leverage the complementary characteristics of the two datasets. StreetGS~\cite{yan2024street} sequences feature challenging background variations but do not contain non-rigid actors, whereas OmniRe~\cite{chen2024omnire} sequences include non-rigid actors while presenting relatively less challenging backgrounds due to the stable ego-view.

\begin{figure*}[ht]
    \centering
    \includegraphics[width=1\linewidth]{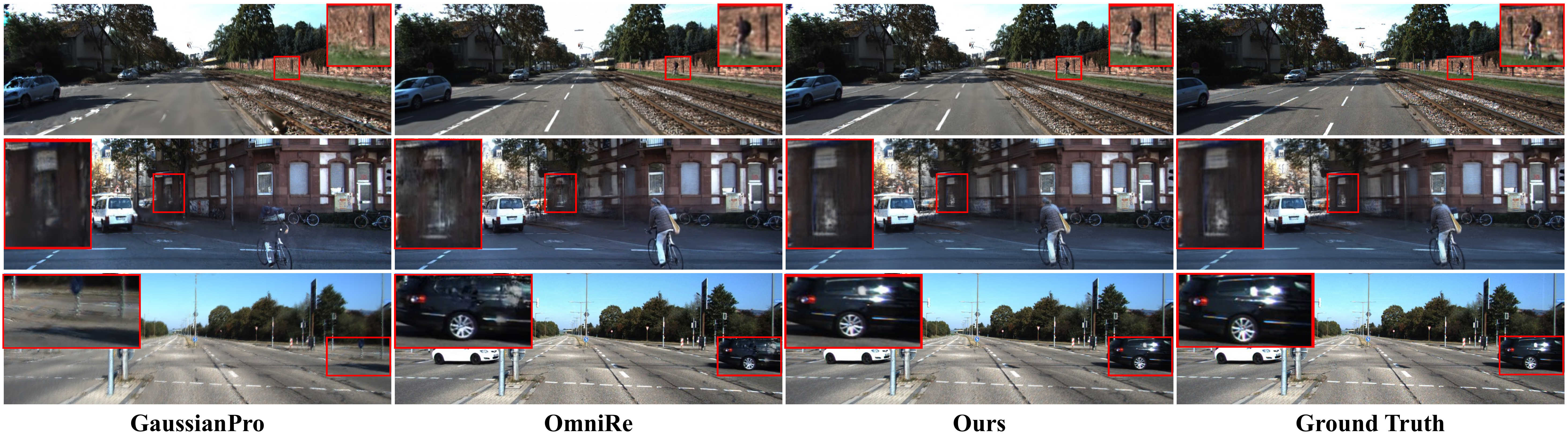}
    \vspace{-0.3cm}
    \caption{\textbf{Qualitative NVS comparison results on the KITTI dataset}. The rendering image resolution is $370 \times 1226$.}
    \label{fig:com_kitti}
\end{figure*}

\noindent \textbf{KITTI}~\cite{geiger2012we}
includes numerous scenes with significant lighting variations, ranging from high-exposure areas to shadowed regions, creating substantial challenges for reconstruction. Therefore, evaluating the reconstruction performance on sequences from the KITTI dataset provides a robustness test of the model's resilience to varying environmental conditions.
We select 3 sequences containing both dynamic non-rigid actors and challenging backgrounds to perform comparative experiments.

\subsection{Implementation Details}
We evaluate the performance of our method in both training-view reconstruction and novel-view synthesis. The novel views are selected by sampling every fourth frame from the original sequence, ensuring that these views are excluded from the model’s training process.
Training a scene of 30K iterations takes 68 minutes on a single NVIDIA L20 GPU, far more efficient than Desire-GS's over 180 minutes (\textcolor{black}{see Table~\ref{tab:ablation_time}}).
We employ DepthAnything-V2~\cite{yang2024depthv2} to generate relative depth priors, ZoeDepth~\cite{bhat2023zoedepth} to obtain metric depth priors, and Omnidata~\cite{eftekhar2021omnidata} to provide surface normal priors.
During point cloud initialization, we employ COLMAP~\cite{schonberger2016structure} to obtain SfM points, which are then fused with LiDAR points.
During training, the gradient information is incorporated after 500 iterations to update the neural Gaussians and adjust the visibility of anchor points. \textcolor{black}{After} 1500 iterations, neural Gaussians are further refined, including the removal of redundant or invisible anchors, addition of new anchors, and updates to properties such as opacity. We train our model with 30,000 iterations on both the Waymo and KITTI datasets.
In our training process, we set the hyperparameters as follows: 
$\phi_1=10, \phi_2=4, \lambda_d=0.01, \lambda_n=0.01, \lambda_m=0.01, \lambda=0.2$.

\subsection{Benchmark Evaluation}

\paragraph{Results on Waymo.}
We evaluate our approach against established methods, including both static methods and dynamic methods, and use both reconstruction and novel view synthesis metrics on the Waymo dataset (Table~\ref{tab:waymo}). Our method surpasses all baselines in both PSNR and LPIPS reconstruction metrics, showcasing high precision. While StreetGS~\cite{yan2024street} and OmniRe~\cite{chen2024omnire} perform well in reconstruction metrics, they struggle with novel view synthesis, highlighting their limitations in handling viewpoint transitions. In contrast, DriveSplat excels in novel view synthesis tasks, outperforming all baselines across three evaluation metrics. Visual analysis (Fig.~\ref{fig:com_waymo}) highlights DriveSplat’s superior artifact-free rendering of vehicles, enhanced clarity of challenging static background details, and accurate depiction of dynamic vehicles.

We compare our method with representative static and dynamic baselines on the Waymo dataset using both reconstruction and novel view synthesis metrics, as reported in Table~\ref{tab:waymo}.
DriveSplat consistently achieves the best performance in terms of SSIM and LPIPS for reconstruction, indicating accurate fitting of observed views.
While recent dynamic methods such as StreetGS~\cite{yan2024street} and OmniRe~\cite{chen2024omnire} attain competitive reconstruction scores, their performance degrades noticeably under novel viewpoint evaluation, reflecting limited robustness to viewpoint changes.
In contrast, DriveSplat demonstrates strong generalization in novel view synthesis, outperforming all baselines across all evaluated metrics.
Qualitative results in Fig.~\ref{fig:com_waymo} further illustrate that DriveSplat produces artifact-free renderings of dynamic vehicles, preserves fine-grained static background structures, and maintains consistent geometry under challenging viewpoint transitions.

\begin{table}[t]
    \centering
    \caption{\textbf{Quantitative comparison results focusing on sequences with predominantly non-rigid actors.} PSNR* denotes the rendering quality at dynamic regions.}
    \label{tab:dynam}
    \setlength{\tabcolsep}{5pt}
    \renewcommand{\arraystretch}{1.0}
    \begin{tabularx}{\linewidth}{@{} lcccc @{}}
    \toprule
        Model & PSNR↑ & SSIM↑ & LPIPS↓ & PSNR*↑\\ 
        \midrule
        Deform-GS~\cite{yang2024deformable} & 28.19 & 0.903 & 0.135 & 22.58\\
        4D-GS~\cite{wu20234d} & 23.77 & 0.860 & 0.121 & 17.90\\
        OmniRe~\cite{chen2024omnire} & \underline{30.69} & \underline{0.906} & \underline{0.107} & \underline{28.79}\\
        Ours & \textbf{30.96} & \textbf{0.913} & \textbf{0.104} & \textbf{29.46}\\
        \bottomrule
    \end{tabularx}
\end{table}

To specifically evaluate the effectiveness of our method in reconstructing non-rigid actors compared to other dynamic reconstruction techniques, we conducted assessments on two select sequences from the Waymo dataset, which contain extensive pedestrian activity. We utilized the PSNR* metric to focus specifically on the performance pertaining to dynamic components, and compare with recent non-rigid optimization methods, including Deform-GS \cite{yang2024deformable}, 4D-GS \cite{wu20234d}, and OmniRe \cite{chen2024omnire}. As shown in Table \ref{tab:dynam}, our method surpasses other reconstruction approaches that incorporate deformation networks.

\paragraph{Results on KITTI.}
The performance of DriveSplat is also evaluated on the KITTI dataset. 
As detailed in Table \ref{tab:kitti}, DriveSplat outperforms baselines in both reconstruction and novel view synthesis tasks. Fig. \ref{fig:com_kitti} highlights rendering results, where DriveSplat demonstrates superior background clarity and accurate rendering of dynamic vehicles. Compared to OmniRe~\cite{chen2024omnire}, our method demonstrates improved performance in novel view synthesis, particularly in preserving background details, as illustrated in the second row of Fig. \ref{fig:com_kitti}.

\subsection{Challenging NVS Comparison}\label{sec:nvs}

\begin{figure}[t]
    \centering
    \includegraphics[width=1.0\linewidth]{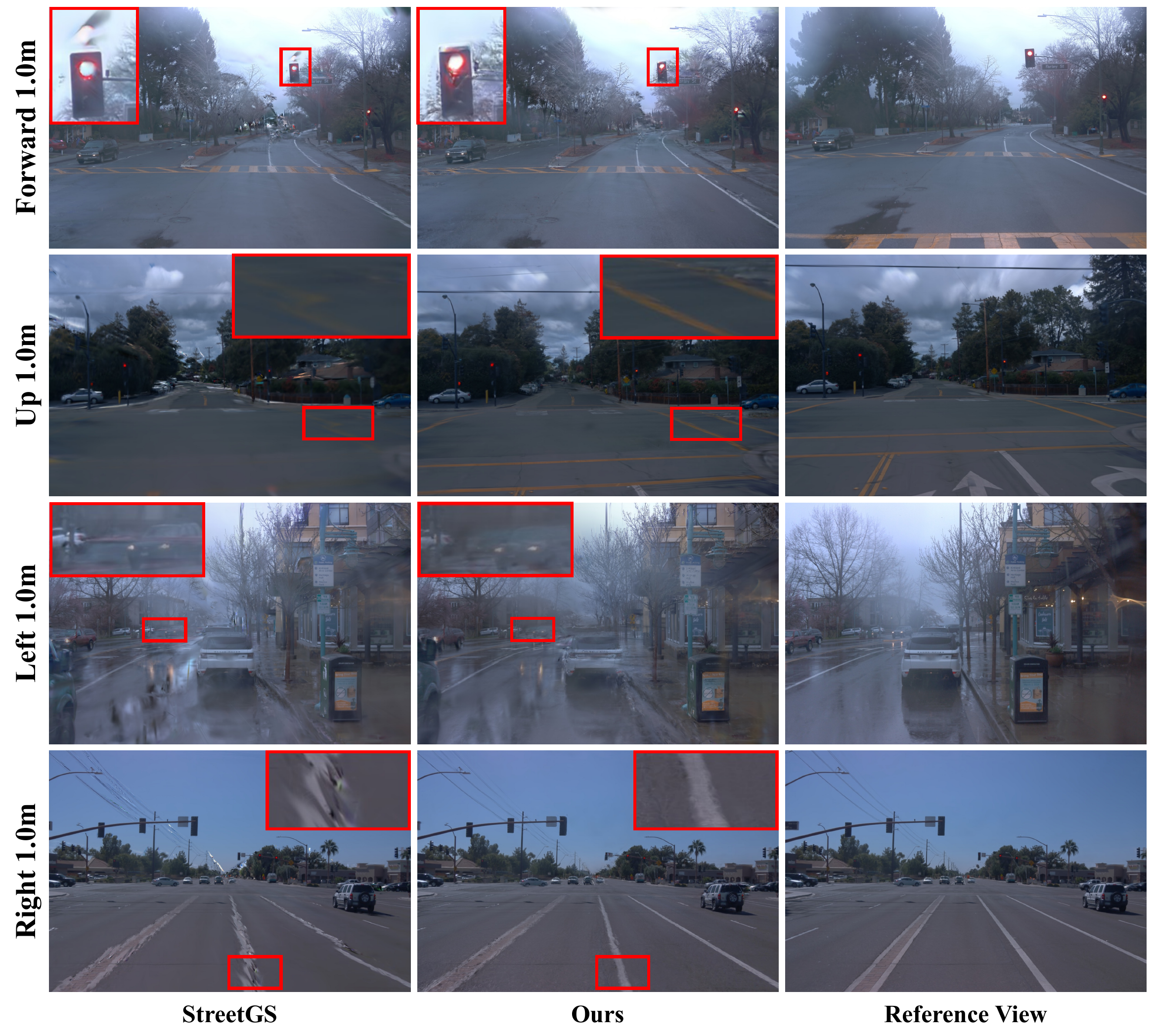}
    \vspace{-0.3cm}
    \caption{\textbf{Qualitative comparison of NVS rendering results on Waymo}. Four challenging novel views are obtained by applying a 1.0-meter shift in viewpoint.}
    \label{fig:com_nvs}
\end{figure}

To evaluate the rendering performance of our method under challenging novel view conditions, we conduct experiments on the Waymo dataset by modifying the original training viewpoints and visualizing the outcomes. Specifically, we apply a 1.0-meter viewpoint shift and compare our method with StreetGS~\cite{yan2024street}, as illustrated in Fig. \ref{fig:com_nvs}. Our method consistently maintains high-quality rendering of static background elements, such as parked vehicles and road surfaces. For instance, under a 1.0-meter rightward viewpoint shift (Fig. \ref{fig:com_nvs}), our approach preserves fine-grained details of road markings, including white lane lines, whereas StreetGS~\cite{yan2024street} exhibits noticeable rendering artifacts, particularly on road markings.

For quantitative evaluation, we conduct a comprehensive comparison under camera viewpoint shifts of 1 m, 2 m, and 5 m, corresponding to increasing degrees of viewpoint change in driving scenarios. As shown in Table~\ref{tab:nvs_com}, our method consistently outperforms competing approaches under small viewpoint shifts (1 m and 2 m), demonstrating robust novel-view synthesis for moderate viewpoint changes commonly encountered in real-world driving scenes. Under the largest shift of 5 m, all methods exhibit comparable performance, reflecting the inherent difficulty of large view extrapolation in driving scenarios when relying solely on reconstruction-based approaches.

\begin{table}[t]
    \centering
    \caption{\textbf{Quantitative evaluation on a challenging novel-view synthesis task}. Results are reported under three camera viewpoint shift settings: 1 m, 2 m, and 5 m.}
    \setlength{\tabcolsep}{2.8pt}
    \renewcommand{\arraystretch}{1.0}
    \begin{tabularx}{\linewidth}{@{} lcccccc @{}}
    \toprule
        \multirow{2}{*}{\textbf{Model}} & \multicolumn{2}{c}{\textbf{1m}} & \multicolumn{2}{c}{\textbf{2m}} & \multicolumn{2}{c}{\textbf{5m}}\\
        \cmidrule(lr){2-3} \cmidrule(lr){4-5} \cmidrule(lr){6-7}
         & FID↓  & FVD↓ & FID↓ & FVD↓ & FID↓ & FVD↓\\ 
        \midrule
        StreetGS~\cite{yan2024street} & 38.3 & 16.8 & 61.9 & 36.7 & \underline{149.6} & \underline{64.9} \\
        OmniRe~\cite{chen2024omnire} & 41.8 & 25.4 & 60.4 & 34.5 & 155.9 & 68.4 \\
        Desire-GS~\cite{peng2025desire}  & 40.8 & 23.6 & 60.5 & 40.4 & 151.6 & 65.8 \\
        AD-GS~\cite{xu2025ad}  & \underline{37.4} & \underline{15.9} & \underline{57.2} & \underline{31.9} & 152.6 & 67.9 \\
        Ours & \textbf{35.8} & \textbf{12.9} & \textbf{50.0} & \textbf{27.2} & \textbf{149.3} & \textbf{64.0} \\
        \bottomrule
    \end{tabularx}
    \label{tab:nvs_com}
\end{table}

\subsection{Time Efficiency Analysis}
\begin{table}[t]
    \centering
    \caption{\textbf{Training and rendering speed comparison}. PSNR shows the rendering result from test views.}
    \setlength{\tabcolsep}{6pt}
    \renewcommand{\arraystretch}{1.0}
    \begin{tabularx}{\linewidth}{@{} lccc @{}}
    \toprule
        Model & Train (min)↓  & FPS↑ & PSNR (dB)↑\\ 
        \midrule
        StreetGS~\cite{yan2024street} & 84.6 & 44.1 & 30.24\\
        OmniRe~\cite{chen2024omnire}  & \textbf{68.2} & 51.5 & 31.19\\
        Desire-GS~\cite{peng2025desire}  & 182.9 & 37.8 & 32.35\\
        AD-GS~\cite{xu2025ad}  & 88.6 & \underline{54.2} & \underline{33.08}\\
        Ours & \underline{69.6} & \textbf{73.5} & \textbf{33.83}\\
        \bottomrule
    \end{tabularx}
    \label{tab:ablation_time}
\end{table}
To evaluate the time efficiency of our method, we conduct a comparative study with several state-of-the-art dynamic--static disentangled driving-scene reconstruction approaches, including StreetGS~\cite{yan2024street}, OmniRe~\cite{chen2024omnire}, Desire-GS~\cite{peng2025desire}, and AD-GS~\cite{xu2025ad}. 
The comparison covers both training efficiency and rendering performance, as well as the corresponding reconstruction quality. 
The quantitative results are summarized in Table~\ref{tab:ablation_time}.
Our model demonstrates a balanced performance with a training time of 69.6 minutes and a rendering speed of 73.5 FPS, outperforming StreetGS~\cite{yan2024street} and AD-GS~\cite{xu2025ad}. 
Desire-GS~\cite{peng2025desire} employs a two-stage training strategy and requires over 50,000 iterations, resulting in a lengthy training process. Additionally, our model achieves a superior rendering quality, indicating better novel-view synthesis quality.

\subsection{Ablation Study} \label{sec:ablation}
\paragraph{Ablation study on initialization module.}
We evaluate the impact of different initialization methods, as shown in Table \ref{tab:ablation3}. 
LiDAR delivers the most precise point clouds but lacks coverage for tall buildings and distant areas. Although SfM provides sparser points, it offers broader scene coverage and thus slightly outperforms LiDAR alone. DUSt3R yields the densest point clouds, but due to misalignment in scale and position with real-world coordinates, even after transformation, its performance is suboptimal. Consequently, we selected the SfM+LiDAR combination for initialization, which produced the best rendering results.

\paragraph{Ablation study on background representation module.}
Our background representation module incorporates three key components: scene-aware multi-scale Gaussian representation (SMG), geometry-guided scene partitioning (GSP), and learnable level-of-detail allocation (LLOD). 
To evaluate the contribution of each component, we perform a comprehensive ablation study, with the results summarized in Table~\ref{tab:ablation4}.
As a baseline, we replace the proposed background representation module with that adopted in StreetGS~\cite{yan2024street}, while keeping all other settings unchanged. 
As shown in Table~\ref{tab:ablation4}, we add these components cumulatively, and each addition consistently improves rendering quality. 
Moreover, SMG and LLOD notably improve rendering efficiency, indicating their effectiveness in reducing computational overhead while preserving reconstruction fidelity.

\begin{figure}[t]
    \centering
    \includegraphics[width=1.0\linewidth]{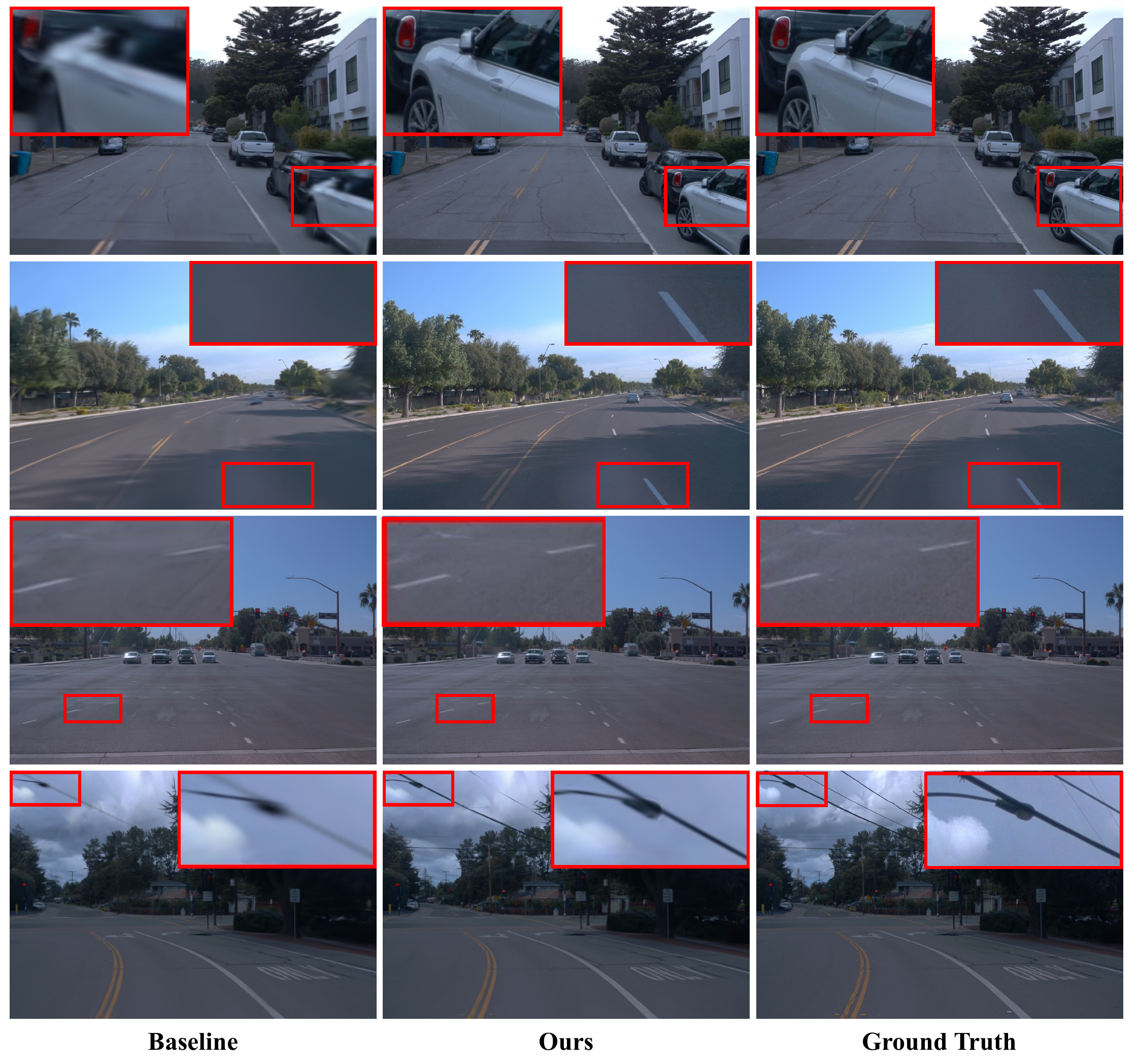}
    \caption{\textbf{Qualitative ablation study of background representation module}. Our method yields improved rendering details compared to the baseline.}
    \label{fig:bkgd_opt}
\end{figure}

Furthermore, we provide more visualization results to show the improvement of our method. As shown in Fig. \ref{fig:bkgd_opt}, the use of our background optimization method can better preserve the background details in close-range regions and reduce the artifacts.

\begin{table}[t]
    \centering
        \centering
        \caption{\textbf{Ablation study of the point cloud initialization module}. Fusion* indicates our final choice of combining SfM and LiDAR data.}
        \setlength{\tabcolsep}{8pt}
        \renewcommand{\arraystretch}{1.0}
        \begin{tabularx}{\linewidth}{@{} lcccc @{}}
        \toprule
            Initialize & PSNR↑ & SSIM↑ & LPIPS↓ & Abs Rel↓\\ 
            \midrule
            Random & 29.64 & 0.894 & 0.127 & 0.269\\
            SfM & 33.16 & 0.912 & 0.112 & 0.239\\
            LiDAR & 32.98 & 0.913 & 0.114 & \textbf{0.174} \\
            DUSt3R & 31.42 & 0.887 & 0.118 & 0.209 \\
            Fusion* & \textbf{33.83} & \textbf{0.923} & \textbf{0.103} & 0.185\\
        \bottomrule
        \end{tabularx}
        \label{tab:ablation3}
\end{table}

\begin{table}[t]
        \centering
        \caption{\textbf{Ablation study on background representation module}. SMG, GSP, and LLOD denote the scene-aware multi-scale Gaussian representation, geometry-guided partitioning, and learnable level-of-detail allocation, respectively.}
        \setlength{\tabcolsep}{8pt}
        \renewcommand{\arraystretch}{1.0}
        \begin{tabularx}{\linewidth}{@{} lcccc @{}}
        \toprule
            Model & PSNR↑ & SSIM↑ & LPIPS↓ & FPS↑ \\ 
            \midrule
            Baseline & 31.98 & 0.915 & 0.119 & 50.87\\
            + SMG & 32.94 & 0.918 & 0.109 & 66.72\\
            + GSP & 33.47 & 0.919 & 0.107 & 61.60 \\
            + LLOD & \textbf{33.83} & \textbf{0.923} & \textbf{0.103} & \textbf{73.52} \\
        \bottomrule
        \end{tabularx}
        \label{tab:ablation4}
\end{table}

\begin{figure}[t]
    \centering
    \includegraphics[width=1.0\linewidth]{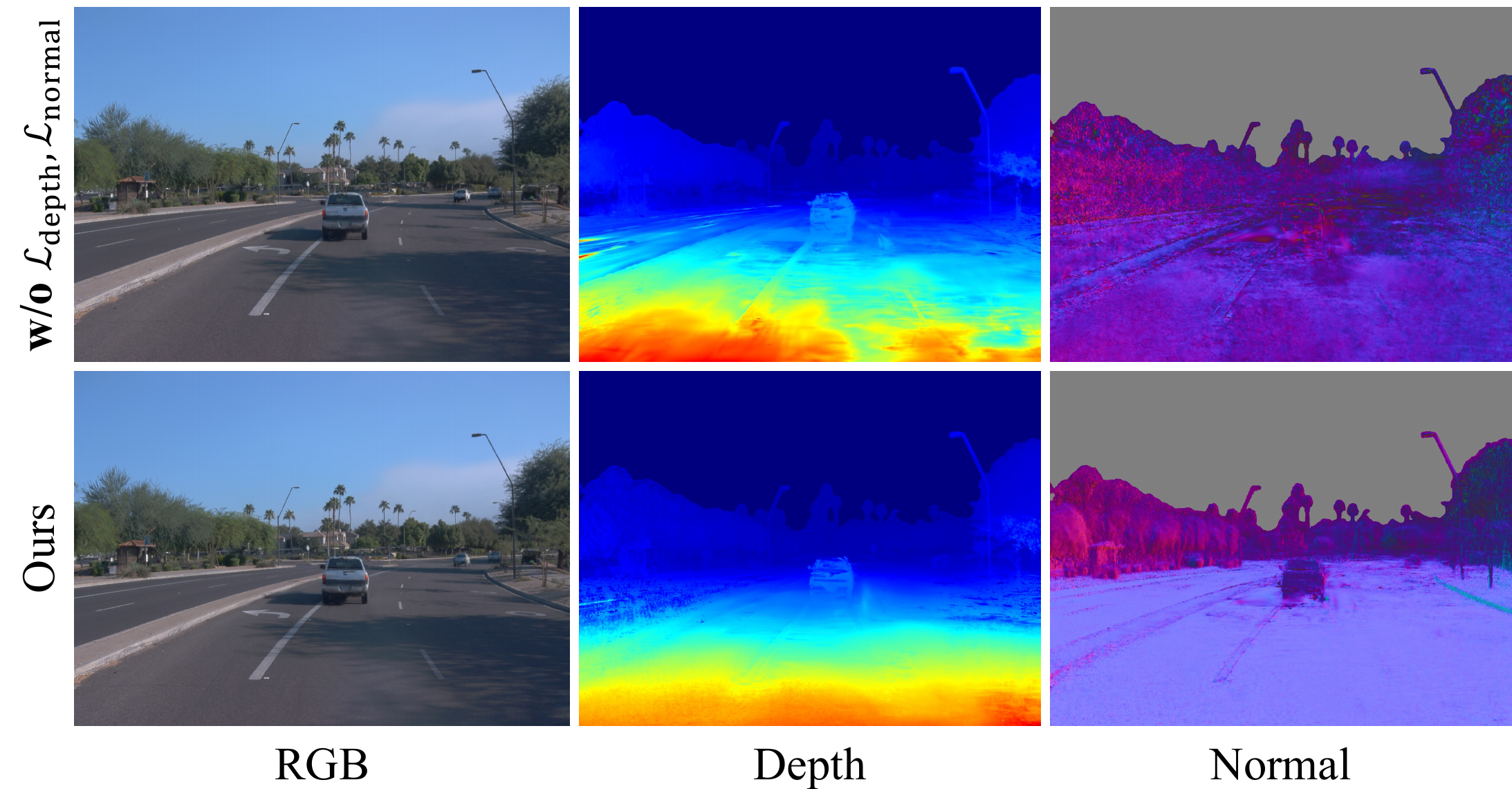}
    \caption{\textbf{Qualitative ablation study results on geometry supervision}.}
    \label{fig:prior}
\end{figure}

\paragraph{Ablation study on geometry optimization.}
We conducted depth and normal rendering experiments using GaussianPro~\cite{cheng2024gaussianpro} as a baseline due to its high geometric accuracy in driving scenarios.
For depth evaluation, we compare our method using depth priors (including metric depth $D_\text{m}$ and relative depth $D_\text{r}$) and without, as shown in Table~\ref{tab:depth_ablation}. 
\begin{table}[t]
    \centering
    \caption{\textbf{Quantitative ablation study results of depth and normal optimization module}.}
    \label{tab:depth_ablation}
    \small                            %
    \setlength{\tabcolsep}{1.0pt}
    \renewcommand{\arraystretch}{1.0}
    \begin{tabularx}{\linewidth}{@{} lcccc @{}}
    \toprule
        Depth Model & AbsRel↓ & MAE↓ & $\delta$1.25↑ & PSNR↑ \\ 
    \midrule
        GaussianPro~\cite{cheng2024gaussianpro} & 0.527 & 9.53 & 56.27 & 31.28 \\
        Ours (w/o $\mathcal{L}_{\text{depth}}$)  & 0.288 & 5.26 & 69.23 & 33.20 \\
        Ours ($D_{\text{m}}$)  & \textbf{0.124} & \textbf{3.41} & \textbf{81.99} & 33.12 \\
        Ours ($D_{\text{r}}$) * & 0.179 & 3.87 & 79.38 & \textbf{33.83} \\
    \midrule
        Normal Model & MAE↓ & RMSE↓ & Simi.↑ & PSNR↑  \\ 
    \midrule
        GaussianPro~\cite{cheng2024gaussianpro} & 1.89 & 2.65 & 0.082 & 31.28 \\
        Ours (w/o $\mathcal{L}_{\text{normal}}$) & 1.21 & 2.24 & 0.276 & 33.31 \\
        Ours (w/ $\mathcal{L}_{\text{normal}}$) * & \textbf{0.84} & \textbf{1.78} & \textbf{0.498} & \textbf{33.83} \\
    \bottomrule
    \end{tabularx}
\end{table}
We use three standard metrics: Absolute Relative Error (Abs Rel), Mean Absolute Error (MAE), and the accuracy under threshold ($\delta<1.25$), together with PSNR to assess the quality of rendered images. Our method surpasses the baseline across all metrics. 
While metric depth $D_\text{m}$ yields the most accurate depth, it reduces rendering quality. Conversely, relative depth $D_\text{r}$ enhances rendering but lessens depth accuracy. This trade-off is due to the limited accuracy of absolute depth data, leading us to prefer relative depth for better overall quality.
For normal evaluation, we assess our method both with and without normal priors, using the estimated normals $N_\text{m}$ as referenced GT. We select three normal evaluation metrics: MAE, Root Mean Square Error (RMSE) and cosine similarity (Simi.).
The result shows that our method outperforms the baseline in all metrics and proves the effectiveness of the normal priors. For qualitative comparison, we show the rendered depth and normal results of our method and those without priors in Fig.~\ref{fig:prior}.

\begin{figure}[t]
    \centering
    \includegraphics[width=1.0\linewidth]{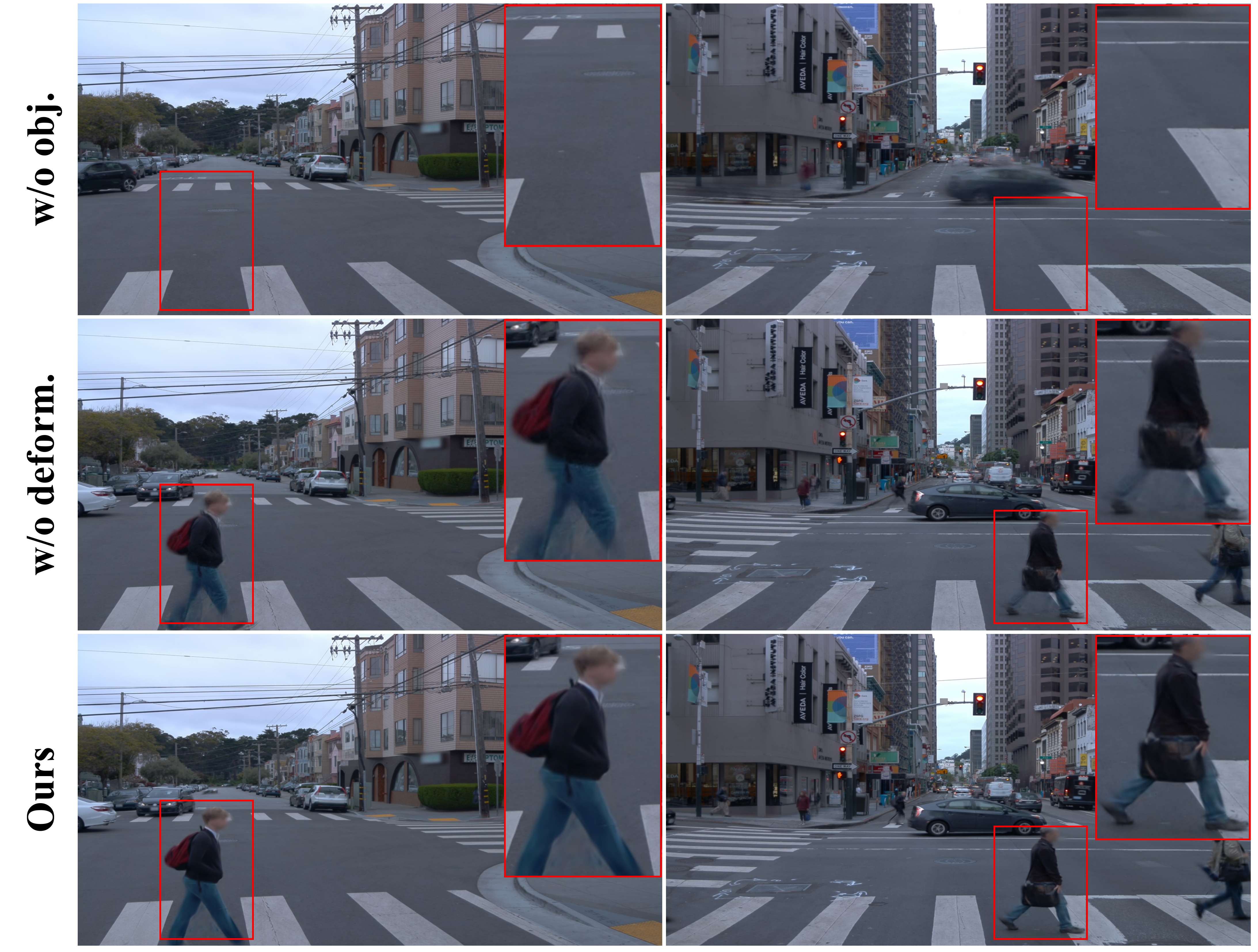}
    \caption{\textbf{Qualitative ablation study results on the dynamic non-rigid actor representation module}.}
    \label{fig:ablation}
\end{figure}

\paragraph{Ablation study on dynamic representation module.}\label{sec:ablation_dynamic}
We utilize four Waymo sequences that contain many pedestrians in front views to evaluate our non-rigid actor reconstruction performance. 
Decoupling dynamic and static elements in the scene effectively improves rendering quality for moving actors.
\begin{table}[t]                   
    \centering
    \caption{\textbf{Quantitative ablation study of the dynamic representation module}. \textit{Deform} denotes the deformable network proposed in Deform-GS~\cite{yang2024deformable}, while deform. denotes our proposed anchor-centric deformable module.}
    \small                            
    \setlength{\tabcolsep}{2pt}
    \renewcommand{\arraystretch}{1.0}
    \begin{tabularx}{\linewidth}{@{} lcccc @{}}
    \toprule
        Model & PSNR↑ & SSIM↑ & LPIPS↓ & FPS↑ \\
    \midrule
        OmniRe~\cite{chen2024omnire} & 30.69 & 0.906 & 0.107 & 37.2 \\
        Ours (w/o deform.) & 28.37 & 0.891 & 0.124 & \textbf{58.9} \\
        Ours (w/ \textit{Deform}) & 30.84 & 0.912 & 0.107 & 39.3\\
        Ours (w/ deform.) * & \textbf{30.96} & \textbf{0.913} & \textbf{0.104} & 48.7 \\ 
    \bottomrule
    \end{tabularx}
    \label{tab:ablation}
\end{table}
Adding individual dynamic object representation can improve the rendering quality of moving vehicles, but it still suffers from motion blur for non-rigid actors. 
After adding the deformable module, the rendering quality of non-rigid actors is significantly improved, as shown in Fig.~\ref{fig:ablation}. Compared to the OmniRe \cite{chen2024omnire} that additionally incorporates SMPL \cite{loper2015smpl}, our method achieves comparable reconstruction performance (Table \ref{tab:ablation}). To evaluate the effectiveness of our proposed non-rigid deformation module, we replace it with the deformable network used in Deform-GS~\cite{yang2024deformable} while keeping all other components unchanged. As shown by the results, our anchor-centric deformation module achieves higher rendering efficiency compared to existing 4D representation methods~\cite{yang2024deformable}. 

\paragraph{Ablation Study of Loss Functions}\label{sec:ablation_loss}

We further verify the effectiveness of the loss functions used in our method, as shown in Table~\ref{tab:ablation_loss}. The ablation study is conducted on four Waymo sequences containing non-rigid actors. These four sequences are also utilized in Table~\ref{tab:ablation}.
All evaluation metrics are obtained from reconstruction experiments using the test views. Among all the loss functions, the $\mathcal{L}_\text{mask}$ performs the most significant improvement on the rendering quality since it can ensure the accuracy of dynamic actors. 

\begin{table}[t]
    \centering
    \caption{\textbf{Ablation study of supervision modules}. Results are derived from reconstruction experiments conducted on the Waymo dataset.}
    \setlength{\tabcolsep}{8pt}
    \renewcommand{\arraystretch}{1}
    \begin{tabularx}{1.0\linewidth}{@{} lccc @{}}
    \toprule
        Model & PSNR↑ & SSIM↑ & LPIPS↓  \\ \midrule
        Ours (w/o $\mathcal{L}_\text{normal}$) & 30.51 & 0.911 & 0.105\\
        Ours (w/o $\mathcal{L}_\text{depth}$) & 30.39 & 0.910 & 0.106\\
        Ours (w/o $\mathcal{L}_\text{mask}$) & 30.14 & 0.908 & 0.108 \\
        Ours & \textbf{30.96} & \textbf{0.913} & \textbf{0.104} \\ 
        \bottomrule
    \end{tabularx}
    \vspace{-0.2cm}
    \label{tab:ablation_loss} 
\end{table}

\section{Application and Discussion}
\subsection{Application of Scene Editing}
\begin{figure}[t]
    \centering
    \includegraphics[width=1\linewidth]{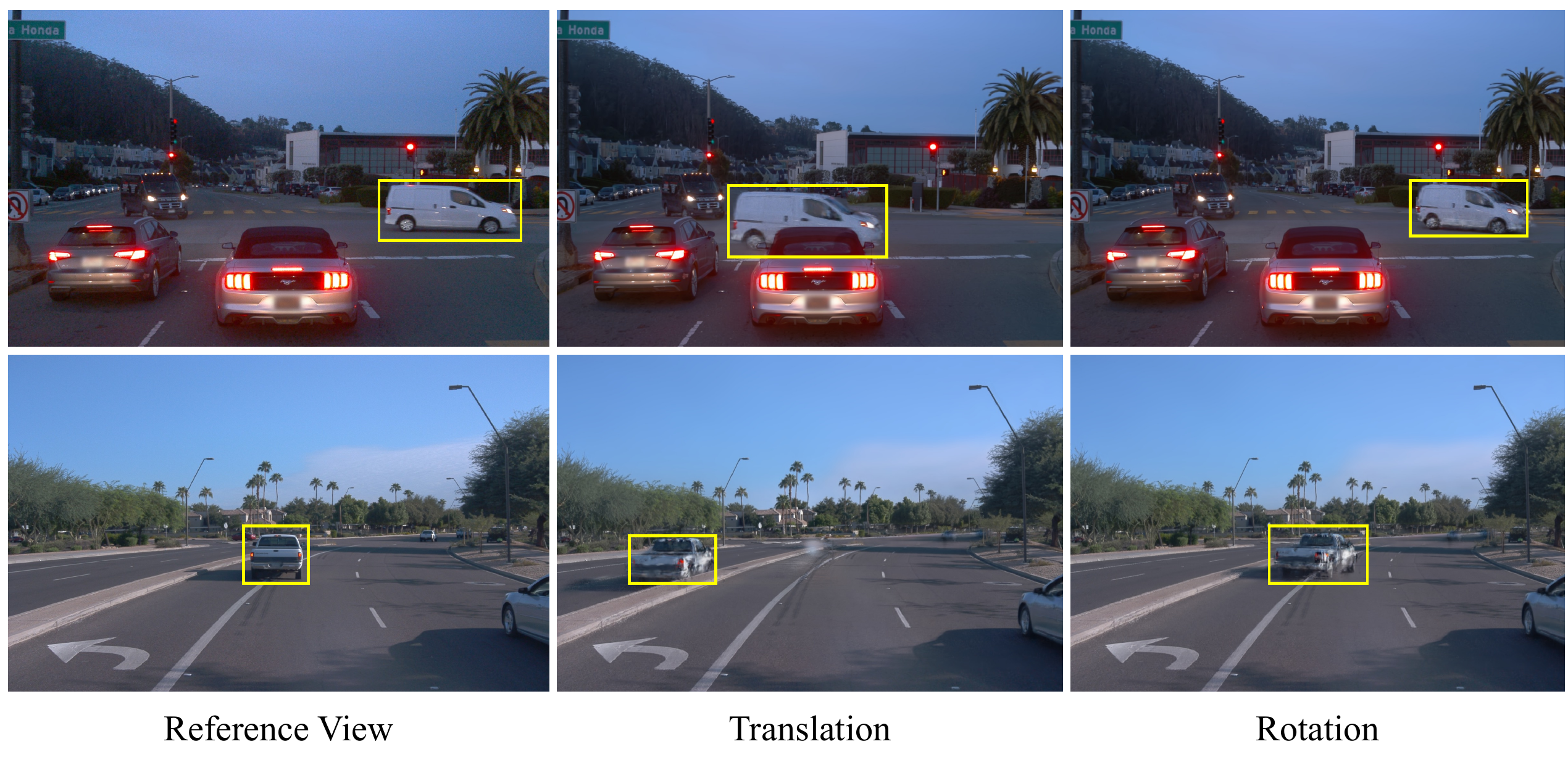}
    \caption{\textbf{Editing operations on the Waymo}. Our method supports dynamic actors editing, including translation and rotation.}
    \label{fig:edit}
\end{figure}

Based on the reconstructed scene, our method enables the editing of foreground dynamic objects, which includes operations such as translation and rotation of specified targets, as illustrated in Fig. \ref{fig:edit}. By allowing precise manipulation of these elements, our approach offers enhanced flexibility and control over scene composition and dynamic interaction.

\subsection{Limitation and Discussion}
Our current approach to 3D reconstruction faces limitations primarily in the domain of scene editing. Specifically, the challenge arises from the insufficient perspective awareness of foreground objects such as vehicles. Due to the lack of full-view perception, these objects cannot be wholly reconstructed, leading to potential deficiencies wherein some aspects of the foreground objects may be missing during editing processes.

To address this limitation, our future work plans to integrate existing advancements in image-to-3D technology. By generating comprehensive 3D representations from single-image foreground inputs, we aim to provide complete 3D assets for editing. This will enable seamless replacement and manipulation using full 3D models, therefore overcoming the present constraint of incomplete object representation during editing.

\begin{figure*}[t]
    \centering
    \includegraphics[width=0.95\textwidth]{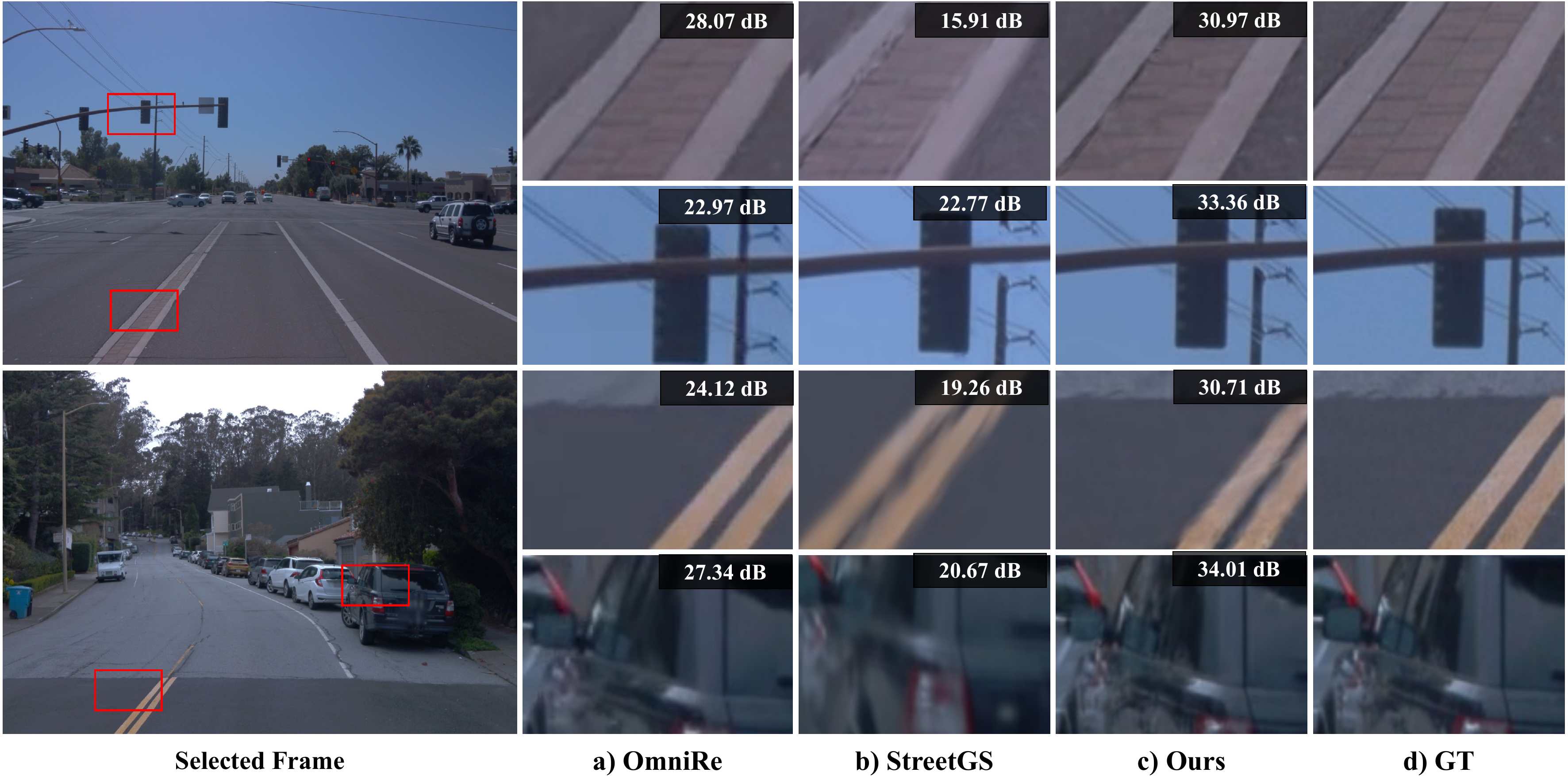}
    \caption{\textbf{Qualitative comparison of detail rendering results on Waymo}. Our method demonstrates superior detail preservation and visual clarity compared to StreetGS~\cite{yan2024street} and OmniRe~\cite{chen2024omnire}, particularly in challenging detailed regions.}
    \label{fig:supp2}
\end{figure*}

However, it is important to note that our current scene editing framework does not yet support this enhancement. We are committed to continuous improvement and anticipate implementing this functionality in subsequent versions of the publicly available code. This effort will ensure that users can leverage fully reconstructed 3D assets in their scene editing endeavors, resulting in more robust and versatile applications.

\section{Conclusion}

We have introduced the DriveSplat, a novel approach for 3D reconstruction in driving scenarios that enhances the accuracy of both static and dynamic elements. By integrating scene-aware multi-scale Gaussian representation with depth and normal priors, our method captures detailed scene geometry for a large-scale background. 
By tracking the poses of moving vehicles and applying anchor-centric motion modeling and deformation propagation to non-rigid actors, dynamic elements achieve accurate and efficient reconstruction. DriveSplat achieves state-of-the-art performance in novel-view synthesis tasks on two autonomous driving datasets, allowing high-quality geometry representation and large-scale scene reconstruction.

\begin{figure*}[t]
    \centering
    \includegraphics[width=0.8\linewidth]{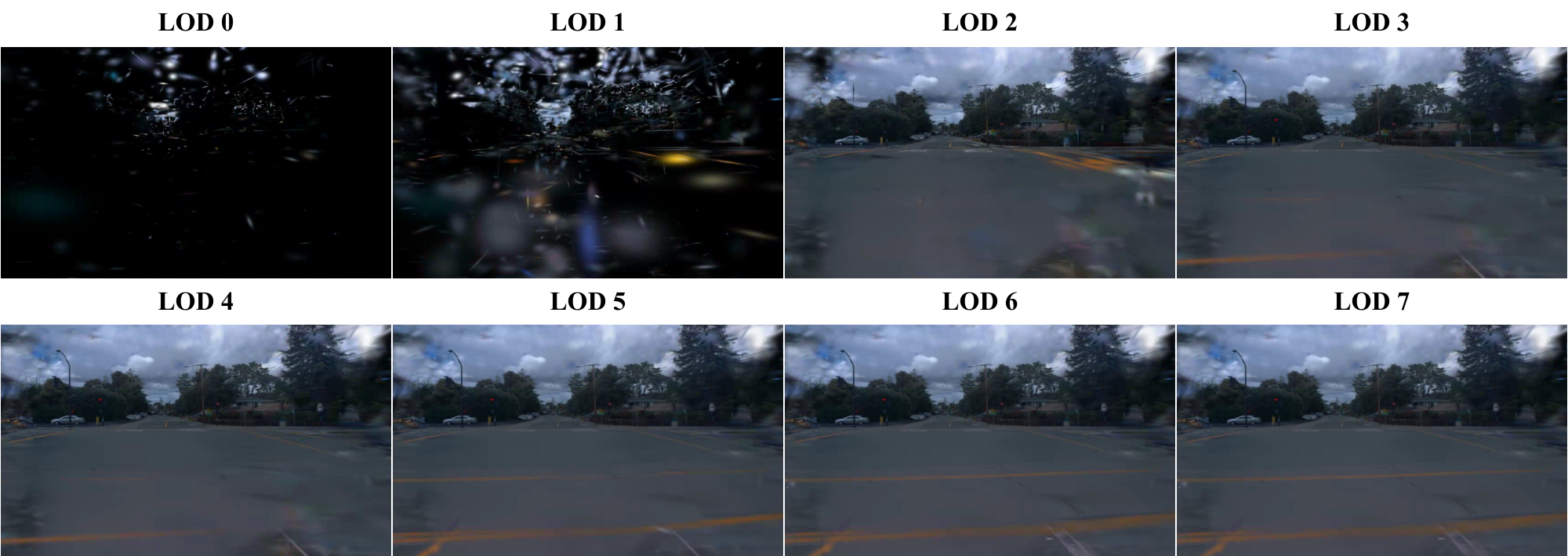}
    \caption{\textbf{Rendering results under different LOD}. Increasing the LOD reveals finer details in near-range regions.}
    \vspace{-0.3cm}
    \label{fig:lod}
\end{figure*}

\begin{figure*}[t]
    \centering
    \includegraphics[width=0.75\linewidth]{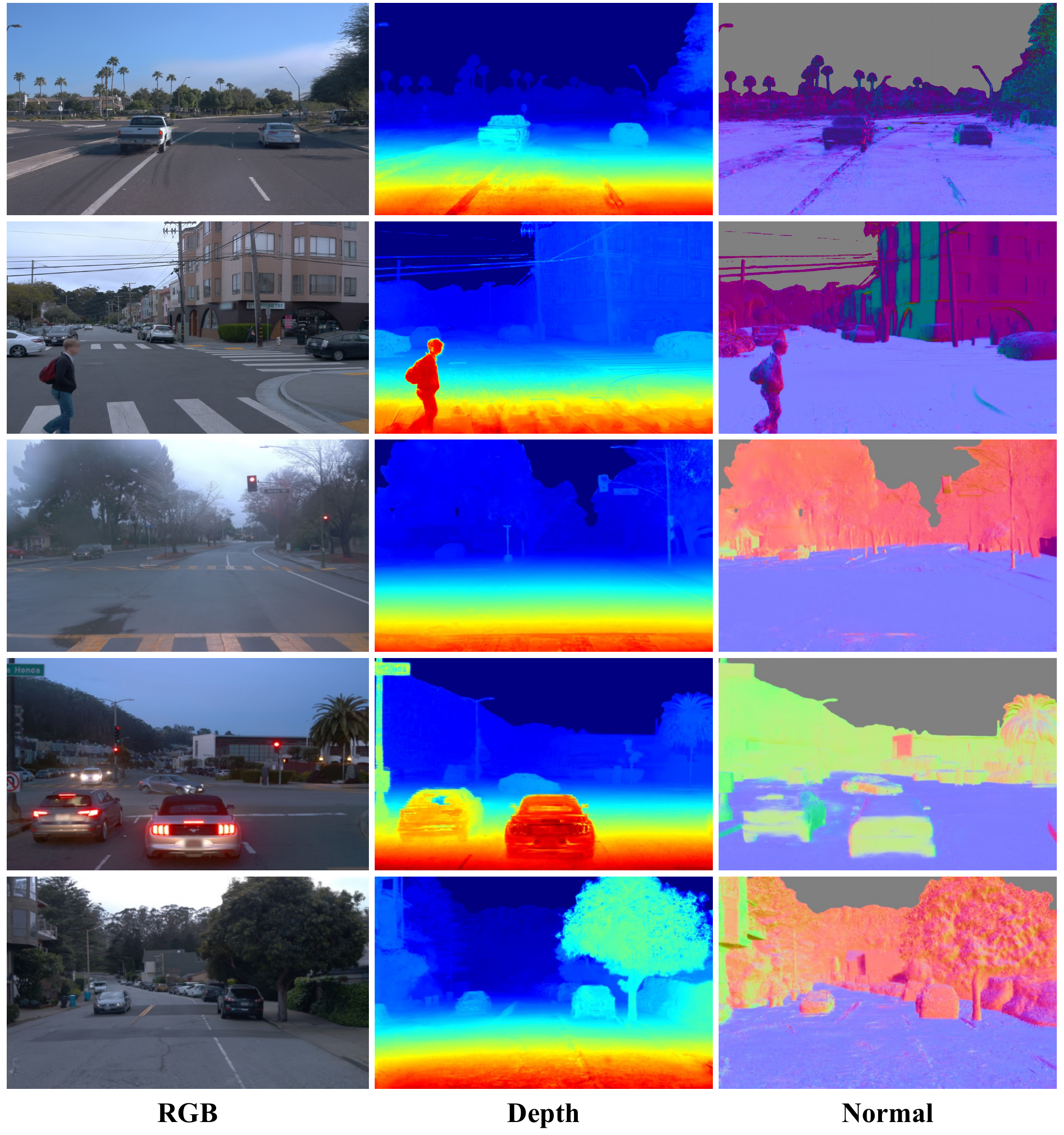}
    \caption{\textbf{Rendering results on the Waymo dataset}. Our method produces high-quality RGB images, depth maps, and normal maps across diverse environmental conditions.}
    \label{fig:render}
\end{figure*}

\section{Appendix}

\subsection{Analysis of LOD}

We adopt a scene-aware multi-scale learnable Gaussian representation for background reconstruction. Our LOD allocation strategy assigns higher levels of detail to near-range regions and lower levels to far-range regions, thereby allocating more representational capacity to visually important areas that occupy a larger portion of the rendered images. As illustrated in Fig.~\ref{fig:lod}, increasing the LOD gradually shifts the rendering focus from far-range to near-range regions. This scene-aware multi-scale representation leads to more detailed rendering results. As shown in Fig.~\ref{fig:supp2}, we compare the rendering quality of fine details and demonstrate that our method outperforms existing approaches both quantitatively and qualitatively.

\subsection{More Visualization Results}\label{sec:results}
Our model generates high-quality RGB images, depth maps, and normal maps, providing a comprehensive representation of the scene for applications like autonomous driving and digital twins. 
The rendering results are shown in Fig.~\ref{fig:render}.
The RGB outputs showcase photorealistic rendering with fine details, the depth maps capture accurate geometric relationships, and the normal maps highlight precise surface orientations, demonstrating the model's versatility and 3D structural accuracy.

\noindent\textbf{Acknowledgements}
This work was supported by the National Natural Science Foundation of China under Grant 62373356 and the Joint Funds of the National Natural Science Foundation of China under U24B20162.

\noindent\textbf{Data Availability}
The Waymo dataset~\cite{sun2020scalability} is publicly available at \url{https://waymo.com/open}.
The KITTI dataset~\cite{geiger2012we} is available at \url{https://www.cvlibs.net/datasets/kitti}.

\noindent\textbf{Conflict of interest.}
The authors have no relevant financial or non-financial interests to disclose.

\bibliography{sn-bibliography}%

\end{document}